\documentclass[review]{elsarticle}

\usepackage{hyperref}
\usepackage{lineno}
\modulolinenumbers[5]
\journal{Computer Speech \& Language}

\usepackage{lipsum}  
\usepackage{xcolor,colortbl}
\definecolor{myrose}{RGB}{248, 229, 216}
\usepackage{float,pdflscape}
\usepackage{lscape}
\usepackage{longtable}
\usepackage{booktabs}
\usepackage{graphicx}
\usepackage{multirow}
\usepackage{rotating}
\usepackage{adjustbox}
\usepackage{caption}
\usepackage{subcaption}
\usepackage{arydshln}
\usepackage[colorinlistoftodos]{todonotes}
\definecolor{darkblue}{rgb}{0, 0, 0.5}
\hypersetup{colorlinks=true,citecolor=darkblue, linkcolor=darkblue, urlcolor=darkblue}
\usepackage{newtxmath}

\usepackage{color} 

\usepackage{titlesec}
\titleformat{\subsubsection}[runin]
  {\normalfont\normalsize\bfseries}{\thesubsubsection}{1em}{}
\titlespacing{\subsubsection}
  {0pt}
  {3.25ex plus 1ex minus .2ex}
  {5pt}

\biboptions{authoryear}

\begin{document}
\begin{frontmatter}
\title{Joint Emotion Label Space Modelling for Affect Lexica}

%
%

\author[mymainaddress]{Luna De Bruyne\corref{mycorrespondingauthor}}
\cortext[mycorrespondingauthor]{Corresponding author.}
\ead{luna.debruyne@ugent.be}

\author[mysecondaryaddress]{Pepa Atanasova}

\author[mysecondaryaddress]{Isabelle Augenstein}

\address[mymainaddress]{Ghent University, Ghent, Belgium}
\address[mysecondaryaddress]{University of Copenhagen, Copenhagen, Denmark}


\begin{abstract}

Emotion lexica are commonly used resources to combat data poverty in automatic emotion detection. However, vocabulary coverage issues, differences in construction method and discrepancies in emotion framework and representation result in a heterogeneous landscape of emotion detection resources, calling for a unified approach to utilising them. To combat this, we present an extended emotion lexicon of 30,273 unique entries, which is a result of merging eight existing emotion lexica by means of a multi-view variational autoencoder (VAE). We showed that a VAE is a valid approach for combining lexica with different label spaces into a joint emotion label space with a chosen number of dimensions, and that these dimensions are still interpretable. We tested the utility of the unified VAE lexicon by employing the lexicon values as features in an emotion detection model. We found that the VAE lexicon outperformed individual lexica, but contrary to our expectations, it did not outperform a naive concatenation of lexica, although it did contribute to the naive concatenation when added as an extra lexicon. Furthermore, using lexicon information as additional features on top of state-of-the-art language models usually resulted in a better performance than when no lexicon information was used.



\end{abstract}


\end{frontmatter}


\section{Introduction}

Affect lexica are valuable resources in the fields of experimental psychology and natural language processing (NLP). An affect lexicon is a list of words or a database that contains lexical entries with their associated affective value. This value can be conveyed as a polarity association (negative - neutral - positive), in which case we call it a sentiment lexicon, or, following emotion frameworks provided in the field of psychology, it can be denoted as a value associated with an emotion category or emotional dimension. In this case, we use the term emotion lexicon.

In the field of psychology, affect lexica are mostly known as affective norms. These norms can be used as stimuli in emotion research or for designing experiments on word memory and processing \citep{Warriner2013}. Also in NLP, emotion lexica are in high demand, because they can be used to combat data poverty in automatic emotion detection. The lexica can be employed as a straight-forward way to automatically label texts with emotional information, or they can be used as features in supervised machine learning approaches \citep{ma-2018-targeted}. Even in state-of-the-art systems for emotion detection (e.g., the winning teams of the SemEval-2018 shared task on multi-label emotion classification), word embeddings in Bi-LSTM architectures are complemented with features from affect lexica \citep{baziotis-etal-2018-ntua, meisheri-dey-2018-tcs}.

However, methodological issues emerge when employing lexica for emotion detection. Firstly, lexica often cover only a small portion of a dataset's vocabulary. Furthermore, the way they are constructed can vary widely: from lab conditions in the field of psychology \citep{Bradley1999}, over crowdsourced annotations \citep{Mohammad2013} to distant supervision \citep{Mohammad2015hashtags}. All these construction methods cause a certain amount of noise, either because of divergence in emotion assessment within or between annotators, or because of imperfections in automatic lexicon creation. Finally, there is currently no consensus on a standard emotion framework: categorical frameworks and dimensional frameworks coexist, in which theorists provide many different sets of categorical labels \citep{ekman1992argument, plutchik1980general} or dimensional axes \citep{mehrabian1974approach, fontaine2007world}. This versatility is also reflected in the existing emotion lexica, which show a myriad of different categorical labels or numerical scales. The inconsistency in labels impedes the exchange of data and knowledge resources and calls for a unified emotion lexicon with a high word coverage that consolidates the annotations of the different emotion frameworks.

Although the problem of vocabulary coverage could be tackled by naively concatenating existing emotion lexica and thus having more entries, this approach does not address the problem of disparate label spaces, nor does it deal with the noise introduced during the construction of the lexica. Naively concatenating different lexica results in conflicting information, which makes the lexicon unsuitable for either research in psychology or keyword-based emotion detection, and could hamper learning in supervised machine learning. By contrast, research showed that merging sentiment lexica by using a multi-view variational autoencoder (VAE) outperforms a naive concatenation technique when the lexicon values are used as features in a supervised learning approach for sentiment analysis  \citep{hoyle-2019-combining}. The intuition of using a VAE to combine lexica, is that the VAE maps the lexica in a shared latent space, making the information in the different lexica less heterogeneous. Moreover, variational autoencoders are commonly used for their noise filtering ability \citep{aggarwal2018neural} and can thus remove the noise introduced in the lexicon construction process.

We believe there is an additional advantage in using a VAE for combining emotion lexica, namely that the dimension of the VAE's latent space can be chosen. While a dimension of three is preferred for sentiment (corresponding to positivity, neutrality and negativity), multiple sizes are possible when dealing with emotions (corresponding to different emotion frameworks). If it shows that the latent dimensions indeed correspond to interpretable emotion dimensions, it opens possibilities for creating large lexica tailored to specific emotion frameworks that are usable in psychology and straightforward keyword-based emotion labeling as well.

However, in comparison to sentiment lexica, joining emotion lexica is more complex: where sentiment lexica contain information about the polarity of words (negative-neutral-positive), emotion lexica contain more fine-grained affective states. This complexity is also reflected in the dimensionality of the lexica and the corresponding emission distributions used in the VAE: where most sentiment lexica are unidimensional, all emotion lexica used in this study are multidimensional. Moreover, different concepts or dimensions are quantified in emotion lexica, e.g. \textit{anger}, \textit{sadness}, \textit{disgust}, \textit{dominance}, etc. This contrasts with sentiment lexica, where the only concept is polarity. The question is thus whether we can still find a meaningful latent space into which the emotion lexica can be mapped.




In this paper, we examine the use of a multi-view variational autoencoder to combine eight existing English emotion lexica in a bigger, joint emotion lexicon and find indications that the chosen dimension of the latent space can be correlated to emotional dimensions present in the source lexica. We then evaluate the joint lexicon on the downstream task of emotion detection on thirteen emotion datasets, by using the lexicon values as features in a logistic/linear regression classifier. We also combine the lexicon features with word embeddings in a Bi-LSTM architecture and show that adding lexicon features improves the performance of plain word embedding models. Contrary to our expectations, the VAE lexicon does not outperform a naive concatenation of lexica, although it does outperform all individual lexica on the task of emotion detection.

\textbf{Contributions}: This paper contributes to the field of emotion analysis in NLP by a) presenting a unified emotion lexicon of 30,273 unique entries, automatically combined by a multi-view variational autoencoder, and show that this 8-dimensional lexicon is still interpretable b) bringing together a large set of existing emotion detection resources and thus learn more about the relationships between them; c) exploring the use of existing lexica and the joint VAE lexicon for the task of emotion detection.

Section \ref{sec:rel_work} describes background on emotion frameworks and related studies that utilise lexica for emotion detection, or that combine lexica and datasets with disparate label spaces. In Section \ref{sec:vae}, we  describe the VAE model (Section \ref{sec:vae_method}) and show how to interpret the dimensions of the resulting joint emotion label space (Section \ref{sec:vae_interpretation}). In Section \ref{sec:evaluation}, we describe our methodology to evaluate the VAE lexicon on the downstream task of emotion detection (Section \ref{sec:experiments}) and report the results (Section \ref{sec:results}), which we further discuss in Section \ref{sec:discussion}. We end this paper with a conclusion in Section \ref{sec:conclusion}.

\section{Related Work}\label{sec:rel_work}
In this section, we will focus on briefly discussing the different frameworks in emotion theory (Section \ref{related_frameworks}), illustrating the use of lexica for emotion detection (Section \ref{related_lexica}) and describing related studies dealing with different emotion frameworks in NLP (Section \ref{related_dealing}).

\subsection{Exploring emotion frameworks} \label{related_frameworks}
Two main approaches of emotion representation exist, namely categorical approaches and dimensional approaches.

In the categorical approach, emotions are represented as specific discrete categories, often with some emotions considered more basic than others. \citeauthor{ekman1992argument}'s \citeyearpar{ekman1992argument} theory of six basic emotions (\textit{joy}, \textit{sadness}, \textit{anger}, \textit{fear}, \textit{disgust}, and \textit{surprise}) is the most well-known, but also \citeauthor{plutchik1980general}'s \citeyearpar{plutchik1980general} wheel of emotions --- in which \textit{joy}, \textit{sadness}, \textit{anger}, \textit{fear}, \textit{disgust}, \textit{surprise}, \textit{trust}, and \textit{anticipation} are considered most basic --- is a common framework in emotion studies. However, many other theorists provide basic emotion frameworks, which can count up to fourteen emotion categories \citep{izard1971face, roseman1984cognitive}.

In the dimensional emotion model, emotions are not seen as discrete emotion categories, but as dimensional concepts. That way, emotions are represented as a point in a multidimensional space. According to \citet{mehrabian1974approach}, every emotional state can be described by scores on the dimensions \textit{valence} (negativity -- positivity), \textit{arousal} (inactivity -- activity) and \textit{dominance} (submission -- dominance), known as the VAD-model. However, in later work, \citet{russell1980circumplex} argued that the two dimensions \textit{valence} and \textit{arousal} suffice for describing emotional states, whereas \citet{fontaine2007world} suggest adding a fourth dimension: \textit{unpredictability}.

In general, categorical representations are easier to interpret compared to dimensional representations. However, categorical frameworks are more restricted, as they are mostly limited to basic emotions, while dimensional representations can describe any affective state.

Various resources have been created based on these different frameworks, including emotion lexica. Most emotion lexica provide numerical values per word, either for the dimensions \textit{valence}, \textit{arousal} and \textit{dominance} \citep{Bradley1999, Warriner2013, Mohammad2018ACL}, or for basic emotions \citep{Stevenson2007, Mohammad2015hashtags, Mohammad2018LREC}. Other lexica just annotate each word with one or more emotion categories \citep{strapparava2004, Mohammad2013}, corresponding to binary tags. The heterogeneity of emotion frameworks is thus reflected in the landscape of emotion lexica (see \ref{appendix:a} for a detailed description of existing emotion lexica).

Also emotion datasets have been produced, in which sentences from different textual genres are annotated with emotional information with the goal of training and testing emotion detection models. Here, the categorical framework clearly dominates, mostly with label sets following Ekman's six \citep{strapparava-mihalcea-2007-semeval, mohammad-2012-emotional, Li2017}, Plutchik's eight \citep{mohammad-2015-electoral, schuff-etal-2017-annotation} or variations thereof \citep{alm-etal-2005-emotions, mohammad-etal-2018-semeval}. Although \citet{strapparava-mihalcea-2007-semeval} approach the Ekman's emotions in a dimensional way (by predicting intensities of emotion categories), the only datasets which truly employ the dimensional emotion model (for English data) are the ones of \citet{Preoctiuc2016} and \citet{Buechel2017}.

\subsection{Using lexica for emotion detection} \label{related_lexica}
Lexica have been the main approach for tackling the task of sentiment analysis (and by extension emotion detection) for a long time \citep{cambria2017practical}. On their own, they can be used to score sentences in a straight-forward way (e.g. by summing scores of sentiment-bearing words in sentences and averaging them), which is the so-called keyword-based approach \citep{ohana2009sentiment}. Moreover, they can serve as features in a supervised learning setting \citep{bravo2014meta}.

In 2007, the first shared task on emotion detection was organised by \citet{strapparava-mihalcea-2007-semeval} as the Affective Text task in the SemEval series. The task was to identify Ekman's emotion categories and valence in news headlines. UPAR \citep{chaumartin-2007-upar7} ended first in the subtask of identifying emotion categories and opted for a keyword-based approach with the sentiment lexicon \texttt{SentiWordNet} \citep{esuli2006sentiwordnet} and the Ekman emotion lexicon \texttt{WordNet Affect} \citep{strapparava2004}. The task organizers themselves experimented with two approaches: one based on the \texttt{WordNet Affect} lexicon and one corpus-based approach, in which they made use of mood-annotated blog posts as training data \citep{strapparava2008learning}. Overall, the organizers' lexicon-based approach gave the best performance.

\citet{chaffar2011using} used \texttt{WordNet Affect} scores as features (together with bag of word  and n-gram features) on the \textsc{Affective Text} \citep{strapparava-mihalcea-2007-semeval}, \textsc{Tales} \citep{alm-etal-2005-emotions} and \textsc{Blogs} \citep{aman-2007-blogs} datasets with Decision Trees, Naive Bayes and SVM as classifiers. Also \citet{kirange2012emotion} performed experiments on the \textsc{Affective Text} dataset and used \texttt{WordNet Affect} lexicon features with an SVM. Indeed, \citet{mohammad-2012-portable} shows that using affect lexica performs better in sentence-level emotion classification than uni- or bigrams alone, using \texttt{WordNet Affect} and the \texttt{NRC Emotion Lexicon} on the \textsc{Affective Text} and \textsc{Blogs} datasets to support this.

Even in recent studies, lexica are still used, for example as features in more sophisticated machine learning systems as deep neural networks. In SemEval-2018 Task 1: Affect in Tweets \citep{mohammad-etal-2018-semeval}, one of the subtasks was a multi-label emotion classification task (with labels \textit{anger}, \textit{anticipation}, \textit{disgust}, \textit{fear}, \textit{joy}, \textit{love}, \textit{optimism}, \textit{pessimism}, \textit{sadness}, \textit{surprise} and \textit{trust}). Apart from word embeddings, emotion and sentiment lexica were the most used features. Even the two best teams \citep{baziotis-etal-2018-ntua, meisheri-dey-2018-tcs} used a Bi-LSTM architecture where word embedding features were complemented with features from affect lexica.

However, relying on lexica to tackle the task of emotion detection has its limitations. The biggest problem is coverage: lexica are often not very extensive. Several studies have tried to expand emotion lexica and used different approaches thereto. \citet{giulianelli2018semi} for example used a label propagation method \citep{zhu2002learning} to expand existing emotion lexica. However, they only work with one original lexicon, namely the \texttt{NRC Emotion Lexicon} \citep{Mohammad2013}. The choice of lexicon was based on the label set: they used the \textsc{Hashtag Emotion Corpus} \citep{Mohammad2015hashtags} which is labeled with the Plutchik emotion categories, and chose their lexicon accordingly. They thus did not have to manage the combination of different label sets, which is another difficulty of using emotion lexica. In the next section, we will discuss some studies that do take into account different emotion frameworks.

\subsection{Dealing with different frameworks} \label{related_dealing}

Due to the sometimes restricted nature of lexica, a unified, expanded emotion lexicon is desirable. This need, however, is complicated by the miscellany of emotion frameworks. To give an example, the word \textit{alien} appears in seven emotion lexica and is thus labeled in seven different ways (see Table \ref{tab:alien}).

\begin{table}
\centering
\small
\begin{tabular}{lcccccccc}
\toprule
\textbf{Lexicon} & \multicolumn{8}{c}{Representations} \\
\midrule
\texttt{Affective Norms}	&	V	&	A	&	D	\\
(1-9 interval)	&	4.45	&	4.86	&	3.56	\\
\\
\texttt{ANEW}	&	V	&	A	&	D	\\
(1-9 interval)	&	5.6	&	5.45	&	4.64	\\
\\
\texttt{NRC VAD}	&	V	&	A	&	D	\\
(0-1 interval)	&	0.41	&	0.615	&	0.491	\\
\\
\texttt{NRC Emotion}	&	Ang	&	Ant	&	Di	&	F	&	J	&	Sa	&	Su	&	T	\\
(binary)	&	0	&	0	&	1	&	1	&	0	&	0	&	0	&	0	\\
\\
\texttt{NRC Affect Intensity}	&	Ang	&	&   &   F	&	J	&	Sa	\\			
(0-1 interval)*	&   -	&	&	&	0.422	&	-	&	-	\\
\\
\texttt{NRC Hashtag}	&	Ang	&	Ant	&	Di	&	F	&	J	&	Sa	&	Su	&	T	\\
(real-valued)*	&	-	&	0.657	&	-	&	0.623	&	-	&	0.640	&	-	&	-	\\
\\
\texttt{Stevenson}	&	Ang	& &	Di	&	F & J & Sa		\\
(1-5 interval)	&	1.47 &	& 1.69 	&	2.42	&	1.29 &	1.28 \\
\bottomrule
\end{tabular}
\caption{Representation of the word \textit{alien} in different lexica. Abbreviations: A = Arousal, Ang = Anger, Ant = Anticipation, D = Dominance, Di = Disgust, F = Fear, J = Joy, Sa = Sadness, Su = Surprise, T = Trust, V = Valence.
\newline
* In some datasets, not all words get a score for each emotion category. In this example, this is indicated with -.}
\label{tab:alien}
\end{table}

\citet{Stevenson2007} and \citet{Buechel2017AFM, buechel-hahn-2018-emotion} investigated mapping methods to shift between categorical and dimensional word representations. This is not only beneficial for lexicon construction, but also for making annotated corpora and tools comparable. In the first study \citep{Stevenson2007}, linear regression was used to predict dimensional (VAD) values from intensity ratings for the categories \textit{happiness}, \textit{anger}, \textit{sadness}, \textit{fear} and \textit{disgust}, and vice versa. They found that no straightforward mapping was possible between intensities of emotion categories and VAD dimensions, but that each emotional category has a different impact on the separate dimensions.

\citet{Buechel2017AFM} trained a kNN model to learn an emotion representation mapping. They used either the intensity ratings of all categories (same as the ones from \citet{Stevenson2007}) to predict one dimension value, or the information of all dimensions to predict the rating of one category. They obtained promising results, with an average Pearson correlation of 0.872 for mapping VAD to an emotion category and 0.844 for mapping categories to dimensions. In subsequent work, a multi-task feed-forward neural network was used to perform the same task and a Pearson correlation of 0.877 was obtained for mapping dimensions to categories and 0.853 for the other direction \citep{buechel-hahn-2018-emotion}. The downside of this approach is that it cannot increase the coverage of a lexicon.

Recently, \citet{buechel-etal-2020-learning} extended their approach so that they could create almost arbitrarily large emotion lexica in any language. The approach combines embedding-based lexicon expansion with emotion representation mapping. Moreover, it exceeds languages by employing a bilingual word translation model. Although this is not exactly a lexicon combination technique, it does allow the creation of a large emotion lexicon with the emotion labels of a desired emotion framework, on the condition that an emotion lexicon with the desired labels exists (even if this lexicon is not in the target language).

Studies that go beyond mappings and actually employ combination techniques are rare on the level of emotion lexicon construction. However, such techniques do exist for sentiment (polarity) lexica. \citet{emerson-declerck-2014-sentimerge} merged four German sentiment lexica by rescaling them linearly (multiplying all the scores by a constant factor per lexicon) and then combining the normalised scores by a Bayesian probabilistic model to calculate latent polarity values, which are assumed to be the `true' values. The original lexica and the merged lexicon all had polarity values on the [-1, 1] interval. The Bayesian model thus just takes care of the noise coming from different sources.

\citet{hoyle-2019-combining} go one step further and combine six lexica with disparate scales, ranging from binary annotations over two-dimensional ratings to 9-point scales. They use a multi-view variational autoencoder to merge the lexica in a latent space of three dimensions. They evaluate these latent scores on nine sentiment analysis datasets and find that they outperform both the individual lexica as well as a naive combination of the lexica.

The problem of different emotion frameworks also emerges when dealing with datasets. \citet{Bostan2018} combine twelve different datasets by means of a rule-based mapping between categorical label sets (e.g. labels like \textit{angry}, \textit{annoyance} and \textit{hate} map to \textit{anger}; \textit{acceptance}, \textit{admiration} and \textit{like} map to \textit{trust}). This results in a final set of eleven emotion categories, in a multi-label approach with continuous values. However, dimensional representations (like the VAD model) are not taken into consideration.




\section{Joint Emotion Space Modelling}\label{sec:vae}

\subsection{Method}\label{sec:vae_method}
There is already a fair number of emotion lexica available for English, however, they all have their own specifics, assets and shortcomings (e.g. regarding emotion framework and vocabulary coverage). Table \ref{tab:lexica} shows an overview of eight emotion lexica with information about their labels and size. More extensive descriptions can be found in \ref{appendix:a}.

\begin{table}
\centering
\small
	\begin{tabular}{ p{2cm} l l r p{3.5cm}}
\toprule
Name & Labels & Values & Size & Reference \\
\hline
\texttt{Affective Norms} & VAD & [1-9]$^3$ & 13,915 & \cite{Warriner2013} \\
\hline
\texttt{ANEW} & VAD & [1-9]$^3$ & 1,034 & \cite{Bradley1999} \\
\hline
\texttt{NRC Emotion} & Plutchik's 8 & \{0, 1\}$^8$ & 14,182 & \cite{Mohammad2013} \\
\hline
\texttt{NRC Affect Intensity} & Ang, F, S, J & [0-1]$^{4}$ & 4,192 & \cite{Mohammad2018LREC} \\
\hline
\texttt{NRC VAD} & VAD & [0-1]$^3$ & 20,007 & \cite{Mohammad2018ACL} \\
\hline
\texttt{NRC Hashtag Emotion} & Plutchik's 8 & [0-$\infty$]$^{8}$ & 16,862 & \cite{Mohammad2015hashtags} \\
\hline
\texttt{Stevenson} & Ang, F, S, J, Di & [1-5]$^5$ & 1,034 & \cite{Stevenson2007} \\
\hline
\texttt{WordNet Affect} & Ekman's 6 & \{0, 1\}$^6$ & 1,113 & \cite{strapparava2004} \\
\bottomrule
\end{tabular}
\caption{Overview of the used emotion lexica. Abbreviations: A = Arousal, Ang = Anger, D = Dominance, Di = Disgust, F = Fear, J = Joy, S = Sadness, V = Valence.
\newline
[ ] = continuous values, \{ \} = discrete values. Exponents refer to the number of dimensions.}
\label{tab:lexica}
\end{table}

For maximum vocabulary coverage, it is appropriate to combine multiple lexica when using lexicon information in an emotion detection task. However, seeing the variety of frameworks and perspectives by which lexica are annotated, this is not self-evident. 

One could say that, when annotating words to create an emotion lexicon $d$, noise is added to the real emotion value $z^w$ of a word $w \in W$, resulting in the observed emotion value $x_d^w$. This noise comes from subjective interpretations, construction method, lab conditions, etc. All emotion values that are observed in a lexicon, are thus distorted. Variational autoencoders (VAEs) are commonly used as noise filters \citep{aggarwal2018neural}, so the idea is that using a VAE, the latent emotion values of each word can be inferred. The noise added by annotation following different frameworks and perspectives, could thus be eliminated using this approach.

\begin{figure}[h!]
    \centering
    \captionsetup{justification=centering}
    \includegraphics[scale=0.4]{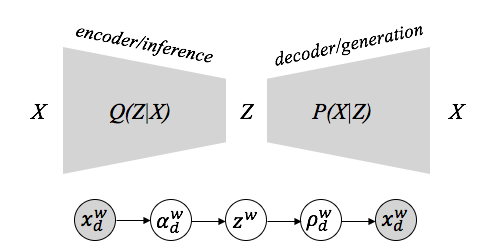}
    \caption{Variational autoencoder model.}
    \label{fig:VAE}
\end{figure}

A traditional VAE consists of an encoder $g$ that takes observed values $X$ as input and outputs parameters for the probability distribution $P(Z|X)$ (which is approximated by a family of distributions $Q_\lambda(Z|X)$), from which we can sample to get a latent representation $Z$. This latent representation is in its turn used as input for a decoder $d$ that outputs the parameters of the probability distribution of the data, in order to reconstruct the original input $X$ (see Figure \ref{fig:VAE}).

Following \citet{hoyle-2019-combining}, and as detailed hereafter, we extend the VAE to a multi-view model, in which each view corresponds to a different lexicon. While a traditional VAE can be employed to build latent representations for words coming from one lexicon and one type of annotation scheme only, a multi-view VAE can learn latent representations of words from different lexica even if they are annotated following different annotation schemes. The multiple possible schemes here represent the multiple views in the VAE. This allows us to join lexica with disparate label spaces, mapping the different labels to a common latent space and resulting in a larger, unified emotion lexicon, which we will call the VAE lexicon.


We use the development sets of thirteen datasets (see Section \ref{sec:evaluation}) to determine the best hyperparameters, i.e. the dimension of the latent variable (search space: 3,6,8,10, 20, 30, 40), the number of nodes in the fully-connected layer of the encoder and decoder network (search space: 82, 128, 256) and the value of the diagonal in the covariance matrices of the emission distributions (see paragraph `Generative network', search space: 0.01, 0.05, 0.1, 0.5). We choose a VAE dimensionality of 8, 82 nodes and a value of 0.05 for the covariance matrices.

\subsubsection*{Inference network}
In the first step, the latent values of $z^w$ are drawn from the prior distribution $P(Z)$, parameterized by $\alpha^w=(\alpha^w_k=1|k \in \{0,..., N\})$, where $N$ is the dimension of the latent variable, and with $\alpha^w_k=1$ we assume a uniform prior for the latent dimensions of each word. The goal of the encoder network or inference network is to find parameters for the posterior distribution $P(Z|X)$ given the prior $P(Z)$ and $P(X|Z)$, where:

\begin{center}
    $P(Z|X) = \frac{P(X|Z)P(Z)}{P(X)}$.
\end{center}

\noindent Calculating $P(X)$ is solved by computing the integral sum:
\begin{equation}
P(X)=\int_{Z} P(X \mid Z) P(Z) d Z
\end{equation}

\noindent However, computing the integral sum over all possible configurations of latent variables is computationally intractable, especially for higher dimensionalities of the latent space, e.g., in this work we experiment with dimensions of up to 8. Therefore, we approximate the posterior distribution with a family of distributions $Q_\lambda(Z)$:

\begin{equation}
Q_{\boldsymbol{\lambda}}(\mathcal{\text{Z}})=\prod_{w \in \mathcal{\text{W}}} Q_{\boldsymbol{\beta}^{w}}\left(\boldsymbol{z}^{w}\right)=\prod_{w \in \mathcal{\text{W}}} \operatorname{Dir}\left(\boldsymbol{\beta}^{w}\right)
\label{eq:posterior}
\end{equation}
\noindent where $\lambda$ are variational parameters.

Whereas the latent variables in regular VAE models are Gaussian, we use a Dirichlet latent variable. A Dirichlet latent variable is preferred as the Dirichlet distribution is a conjugate prior to the multinomial and the categorical distributions. Its posterior being also a Dirichlet distribution, allows for updates of the latent variable from new observations using a closed-form expression. 

We calculate $\beta^{w} \in R^{|Z|}$ as follows: we firstly produce lexicon-specific representations $\boldsymbol{\omega}_{d}^{w} \in R^{|Z|}$ by using an encoder $e$ with lexicon-specific parameters $\phi_{d}$ (each with a dimension of 82) over the emotion values $x_{d}^{w}$ (see Equation~\ref{eq:omega}), and then accumulate these lexicon-specific encodings across lexica according to Equation~\ref{eq:beta}. The Dirichlets are thus more concentrated for words appearing in more lexica.


\begin{equation}
\boldsymbol{\omega}_{d}^{w}=\textit{softmax}\left(e\left(\boldsymbol{x}_{d}^{w} ; \phi_{d}\right)\right)
\label{eq:omega}
\end{equation}

\begin{equation}
\boldsymbol{\beta}^{w}=1+\sum_{d \in \mathcal{D}} \boldsymbol{\omega}_{d}^{w}
\label{eq:beta}
\end{equation}

Note that lexical ambiguities like homographs (words with same spelling but different meanings) are not taken into account while merging the lexica, as there is no information about context or part of speech in the lexica (except for \texttt{WordNet Affect}). Such ambiguous cases may thus have an undesirable effect on the Dirichlets.

\subsubsection*{Generative network}\label{subsec:generative_network}
In the decoder or generative network, $X$ is reconstructed by outputting the likelihood of $X$ given the latent representation $Z$. The joint probability distribution of the data and likelihood is defined as $P(X,Z) = P(X|Z)P(Z)$, where the distribution of the likelihood depends on the lexicon $d$. The decoder outputs parameters for the emission distribution $P(X|Z)$ of the data, from which $X$ is reconstructed. This distribution is lexicon-dependent and is chosen to correspond to the label type in the individual lexicon as explained next. For \texttt{ANEW},  \texttt{Affective Norms} and \texttt{NRC VAD}, which all have three continuous labels (see Table~\ref{tab:lexica}), we model the emission distributions with three-dimensional Gaussians with means $\rho_d^w$ and diagonal covariance matrices equal to 0.05\textit{\textbf{I}}. \texttt{NRC Affect Intensity}, \texttt{NRC Hashtag} and \texttt{Stevenson}, which provide four, eight and five continuous labels respectively, we choose four, eight and five-dimensional Gaussians as emission distributions respectively, with means $\rho_d^w$ and diagonal covariance matrices equal to 0.05\textit{\textbf{I}}. Finally, \texttt{WordNet Affect} and \texttt{NRC Emotion}, where the emotion labels are provided as six and eight binary labels respectively, we choose emission distributions of  six and eight Bernoulli distributions parameterized collectively by $\rho_{d}^{w}$.

First, for each word $w$, a latent vector $z^{w}$ is generated by sampling from the distribution described by the Dirichlet parameters (outputted by the inference network) (Eq.~\ref{latent_z}). For this sampling process, the generalized reparamaterization trick of \citet{ruiz2016generalized} is used. Then, the decoder network $g$ (again a 82-dimensional fully-connected layer) with lexicon-specific weights  $\theta_{d}$ transforms the generated latent emotion value $z^{w}$ into a lexicon-specific representation $\rho^{w}_{d}$ (Eq.~\ref{hidden_zd}). The dimension of $\rho^{w}_{d}$ is lexicon-specific. Finally, the lexicon-specific emotion value is reconstructed from $\rho^{w}_{d}$ with the lexicon-specific emission distribution $P_{d}$ (Eq.~\ref{xw}).
\begin{equation}
    z^{w} \sim Dir(\alpha^{w})
~\label{latent_z}
\end{equation}
\begin{equation}
  \rho^{w}_{d} = g(z^{w}, \theta_{d})
~\label{hidden_zd}
\end{equation}
\begin{equation}
  x_{d}^{w} = P_{d}(x_{d}^{w},\rho^{w}_{d})
~\label{xw}
\end{equation}


\subsection{Interpretation}\label{sec:vae_interpretation}

A VAE allows us to create an extended, unified lexicon by mapping lexica with different label spaces into a common latent space and reducing noise that was introduced during the construction of source lexica. Moreover, another advantage is that the dimensionality of the latent space can be chosen. If we can correlate the dimensions of the latent space to an emotion framework, this opens possibilities to flexibly map different lexica to a target emotion framework and use the resulting lexicon in psychological research and keyword-based emotion detection.

The three most popular emotion frameworks are the basic emotion models by \cite{ekman1992argument} and \cite{plutchik1980general}, consisting of respectively six and eight emotion categories, and the VAD model by \cite{mehrabian1974approach}, consisting of the three dimensions \textit{valence}, \textit{arousal} and \textit{dominance}. Motivated by these frameworks, we included these sizes ($N=3; N=6; N=8$) when testing different dimensionalities for the VAE lexicon.

To test whether the VAE dimensions can be linked to these frameworks, we calculate Spearman's correlation between the dimensions of the VAE lexicon and the emotional dimensions in the source lexica. For measuring correlation with the VAD model, we use \texttt{ANEW}, \texttt{Affective Norms} and \texttt{NRC VAD}. For measuring the correlation with Plutchik, we use \texttt{NRC Hashtag Emotion Lexicon}. Although \texttt{NRC Emotion} is annotated with Plutchik emotions as well, we do not use it in the correlation analysis, because the labels in this lexicon are binary instead of real-valued. For the same reason, we do not correlate the VAE dimensions to the Ekman emotions (as only \texttt{WordNet Affect} is annotated with Ekman emotions and the labels are binary as well).

For VAD, we take all words that are shared between the VAE lexicon (version with $N=3$) and \texttt{ANEW}, \texttt{Affective Norms} and \texttt{NRC VAD},  resulting in 1,034 words (corresponding to the size of the smallest lexicon included in this analysis: \texttt{ANEW}). We determine Spearman's $r$ between the VAE values of all these words and the values in \texttt{ANEW}, \texttt{Affective Norms} and \texttt{NRC VAD} respectively. We do the same for the words in \texttt{NRC Hashtag Emotion Lexicon} (16,862 words) to determine the correlation between the VAE dimensions (version with $N=8$) with the Plutchik categories.

\begin{figure}
    \centering
    \begin{subfigure}[t]{0.9\textwidth}
    \includegraphics[scale=0.6]{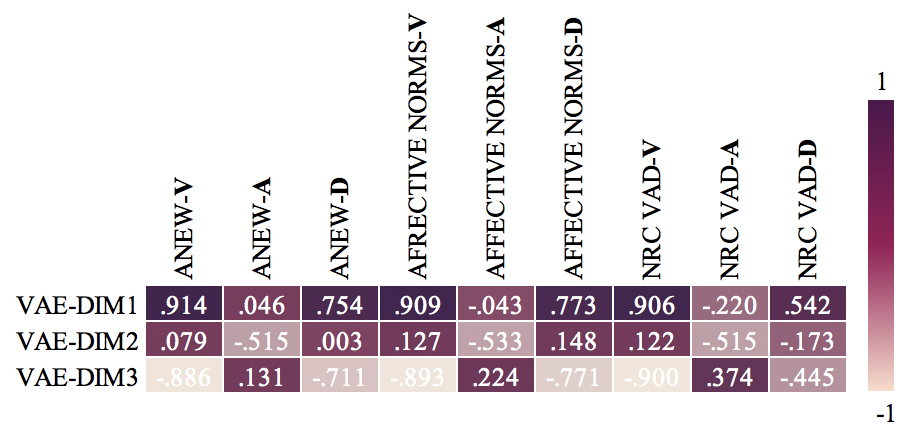}
    \caption{3-dimensional VAE vs. \textit{valance} (\textbf{V}), \textit{arousal} (\textbf{A}) and \textit{dominance} (\textbf{D}) from \texttt{ANEW}, \texttt{Affective Norms} and \texttt{NRC VAD}.}
    \label{fig:vae3_corr}
    \end{subfigure}
    \hspace*{\fill}
    \begin{subfigure}[t]{0.9\textwidth}
    \includegraphics[scale=0.6]{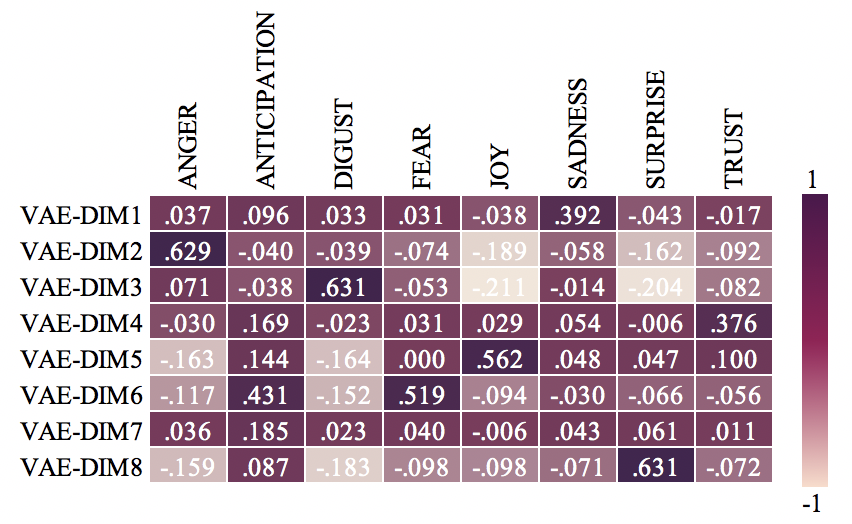}
    \caption{8-dimensional VAE vs. \textit{anger}, \textit{anticipation}, \textit{disgust}, \textit{fear}, \textit{joy}, \textit{sadness}, \textit{surprise} and \textit{trust} from \texttt{NRC Hashtag Emotion}.}
    \label{fig:vae8_corr}
    \end{subfigure}
    \caption{Correlation (Spearman's $r$) between the VAE dimensions and lexicon labels  for their shared words.}
\end{figure}

Correlation coefficients for the VAE version with $N=3$ are shown in Figure \ref{fig:vae3_corr}. We find that there is high correlation ($r > 0.9$) between the first dimension of the VAE lexicon and \textit{valence} and a lower but still strong correlation between this dimension and \textit{dominance} ($0.5 < r < 0.8$). This is in line with the observation of \cite{Warriner2013} that there is a high positive correlation between \textit{valence} and \textit{dominance}. The second dimension has the strongest correlation with \textit{arousal}, although this correlation is negative ($r \approx -0.5$). This means that a high value for this dimension could correspond with a low value for \textit{arousal}. Dimension 3 shows a similar pattern to the first dimension (high correlation with \textit{valence} and \textit{dominance}), although the correlation is negative here. It thus seems that dimension 1 and dimension 3 (when inverted) correspond to valence and dominance (or the other way around) and that arousal is encoded in dimension 2 by means of a negative correlation.
Table \ref{tab:vae_interpretation} shows a collection of words from the VAE lexicon, including `excited', `depressed' and `snake'. The values are in line with the correlation insights. \textit{Valence} and \textit{dominance} could be interpreted as the first and third dimension (where the third dimension should be inverted), although we can not determine which one is which, and the second dimension seems to correspond to inverted arousal. 


\begin{table}
\centering
\small
\begin{tabular}{l|ccc|cccccccc}
 & \multicolumn{3}{c|}{3-dimensional VAE} & \multicolumn{8}{c}{8-dimensional VAE} \\
Word & D1 & D2 & D3 & D1 & D2 & D3 & D4 & D5 & D6 & D7 & D8 \\
\toprule
excited & 5.72 & 1.27 & 1.01 & 1.02 & 1.03 & 1.00 & 1.21 & 4.36 & 2.08 & 1.12 & 1.18 \\
depressed & 1.06 & 4.56 & 5.38 & 4.73 & 1.17 & 4.37 & 1.20 & 1.01 & 1.19 & 1.22 & 1.13\\
furious & 1.16 & 1.12 & 6.72 & 1.09 & 4.60 & 2.87 & 1.12 & 1.00 & 1.10 & 1.14 & 1.09\\
gloomy & 1.04 & 4.91 & 3.05 & 4.95 & 1.24 & 1.92 & 1.22 & 1.01 & 1.20 & 1.29 & 1.17\\
relaxed & 6.76 & 1.96 & 1.28 & 1.50 & 1.22 & 1.08 & 1.56 & 5.44 & 1.23 & 1.56 & 1.41\\
snake & 1.73 & 1.27 & 7.00 & 1.23 & 1.22 & 2.71 & 1.29 & 1.02 & 4.99 & 1.34 & 1.20 \\
tired & 1.40 & 2.10 & 3.50 & 1.78 & 2.36 & 1.53 & 1.32 & 1.05 & 1.30 & 1.36 & 1.30\\
\bottomrule
\end{tabular}
\caption{Examples of some words in the VAE lexicon (version with $N=3$ and $N=8$) with their values per dimension.}
\label{tab:vae_interpretation}
\end{table}

When comparing the VAE version with $N=8$ with the Plutchik emotions, we take the \texttt{NRC Hashtag Emotion Lexicon} as reference. We first have a look at the values of some words (see Table \ref{tab:vae_interpretation}). The word `snake', for example, gets all zero values in the \texttt{NRC Hashtag Emotion Lexicon} except $0.49$ for \textit{disgust} and $0.67$ for \textit{fear}, and has the following scores in the VAE(8) lexicon: $1.23, 1.22, 2.71, 1.29, 1.02, 4.99, 1.34, 1.20$. There are two values standing out (the third and sixth), possibly matching with \textit{disgust} and \textit{fear}.

Indeed, correlation analysis between the VAE dimensions and the categories from \textsc{NRC Hashtag} shows that each of the VAE dimensions has a category which it (highly) correlates with (see Figure \ref{fig:vae8_corr}). The first dimension can be linked to \textit{sadness} ($r \approx 0.4$), the second to \textit{anger} ($r \approx 0.6$), the third to \textit{disgust} ($r \approx 0.6$), the fourth to \textit{trust} ($r \approx 0.4$) and the fifth to \textit{joy} ($r \approx 0.6$). The sixth dimension has the highest correlation with \textit{fear} ($r \approx 0.5$), although it is correlated with \textit{anticipation} as well ($r \approx 0.4$). Also the seventh dimension can be linked to \textit{anticipation}, although the correlation is rather low ($r \approx 0.2$). The eighth dimension is correlated with \textit{surprise} ($r \approx 0.6$). This also corresponds to the `snake' example, which suggested that the third and sixth dimension could be interpreted as \textit{disgust} and \textit{fear}.

The link between VAE dimensions and Plutchik categories is thus rather clear, with only for the seventh dimension a notable ambiguity. Although \textit{anticipation} is the Plutchik category which this dimension correlates with most highly, the correlation is rather low. Moreover, the opposite direction shows something else: when looking at the VAE dimensions which \textit{anticipation} correlates most highly with, the sixth dimension (which we linked to \textit{fear}) comes out as most related. It is thus not evident to say that dimension 7 needs to be interpreted as anticipation. Furthermore, if anticipation is encoded in the VAE dimensions, it is probably not only encoded through dimension 7, but through dimension 6 as well (see also the example of `excited' in Table \ref{tab:vae_interpretation}, which has a high score for dimension 6).

Overall, this correlation analysis suggests that the learned VAE dimensions can be linked to emotional dimensions present in the source lexica and that a joint VAE lexicon can still be interpretable when the dimension is chosen carefully.



\section{Evaluation}\label{sec:evaluation}
\subsection{Method}\label{sec:experiments}

\subsubsection*{Datasets}
We evaluate the VAE joint emotion lexicon by using it as features on the downstream task of emotion detection and compare it with the use of the individual lexica and a naive concatenation thereof. The evaluation is done on eleven commonly used emotion datasets (\textsc{Blogs}, \textsc{Emotion in Text}, \textsc{ElectoralTweets}, \textsc{ISEAR}, \textsc{Tales}, \textsc{TEC}, \textsc{Affect in Tweets}, \textsc{SSEC}, \textsc{Affective Text}, \textsc{EmoBank} and \textsc{Facebook-VA}) and two additional datasets also suited for emotion detection (\textsc{DailyDialog} and \textsc{Emotion-Stimulus}). Information about the labels and size of the datasets is given in Table \ref{tab:datasets} and extra information is given in \ref{appendix:b}.

Eight out of thirteen datasets are annotated in a single-label multi-class categorical approach (each instance has one out of multiple classes as label) and are used for emotion classification. For these datasets, ordinary accuracy (percentage of correct predictions) is our evaluation metric of interest. Two datasets have a multi-label multi-class setup, meaning that each instance can have multiple classes as labels. One instance can thus both have \textit{joy} and \textit{surprise} as labels, which is not possible in the single-label datasets (there it would be \textit{joy} or \textit{surprise} instead of both). Multi-label classification can therefore be seen as the aggregation of multiple binary classification subtasks. We report Jaccard accuracy, a metric specifically used in multi-label tasks and defined as the size of the intersection of the predicted labels and gold labels, divided by the size of their union. Lastly, three datasets have dimensional annotations and are used in a regression task where we calculate Pearson correlations to measure the agreement between gold standard and predicted scores. The Pearson correlations are averaged over the number of dimensions in the particular dataset (6 dimensions in \textsc{Affective Text}, 3 in \textsc{EmoBank} and 2 in \textsc{Facebook-VA}). If a train-test split is provided in the original dataset, we use this. Otherwise, we create an 80:20 train-test split. 10\% from all data in the training set is reserved for development.

\begin{table}
\centering
\footnotesize
	\begin{tabular}{ c p{1.8cm} p{2cm} r p{3.5cm}}
\toprule
Type & Name & Labels & Size & Reference \\
\midrule
\multirow{15}{*}{SL} & \textsc{Blogs} & Ekman, neutral & 4,090 & \cite{aman-2007-blogs} \\
\cline{2-5}
 & \textsc{Emotion in Text} & 13 categories & 40,000 & CrowdFlower \\
\cline{2-5}
 & \textsc{DailyDialog} & Ekman, neutral & 13,118 & \cite{Li2017} \\
\cline{2-5}
 & \textsc{Electoral Tweets} & 19 categories & 4,058 & \cite{mohammad-2015-electoral} \\
\cline{2-5}
 & \textsc{Emotion-Stimulus} & Ekman + Sh & 2,414 & \cite{ghazi2015} \\
\cline{2-5}
 & \textsc{ISEAR}  & Ang, D, F, G, J, Sa, Sh & 7,665 & \cite{scherer1994}\\
\cline{2-5}
 & \textsc{Tales} & Ang, F, J, Sa, Su & 1,207 & \cite{alm-etal-2005-emotions} \\
\cline{2-5}
 & \textsc{TEC} & Ekman & 21,051 & \cite{mohammad-2012-emotional} \\
\midrule
\multirow{3}{*}{ML} & \textsc{Affect in Tweets} & Plutchik + L, O, P & 10,983 & \cite{mohammad-etal-2018-semeval} \\
\cline{2-5}
& \textsc{SSEC} &  Plutchik & 4,868 & \cite{schuff-etal-2017-annotation} \\
\midrule
\multirow{5}{*}{Reg} & \textsc{Affective Text} & Ekman, V (\{0, 1, ... 100\}) & 1,250 & \cite{strapparava-mihalcea-2007-semeval} \\
\cline{2-5}
& \textsc{EmoBank} & VAD ([1-5]) & 10,548 & \cite{Buechel2017} \\
\cline{2-5}
& \textsc{Facebook-VA} & V, A (\{1, 2, ... 9\}) & 2,895 & \cite{Preoctiuc2016} \\
\bottomrule
\end{tabular}
\caption{Overview of the used emotion datasets.
\newline
Abbreviations: A = Arousal, Ang = Anger, D = Disgust, F = Fear, J = Joy, G = Guilt, L = Love, ML = Multi-Label, O = Optimism, P = Pessimism, Reg = Regression, Sa = Sadness, Sh = Shame, SL = Single-Label, Su = Surprise, V = Valence.}
\label{tab:datasets}
\end{table}

\subsubsection*{Machine learning models}
Traditional machine learning and neural methods are used to assess the use of lexica as features for emotion detection. As traditional machine learning approach, we use a logistic regression classifier for the categorical datasets (for the multi-label datasets, we build separate binary classifiers for each of the categories and join the predictions afterwards) and linear regression for the continuous datasets (again, we build separate models for each of the dimensions). The logistic regression classifier uses a liblinear solver with L2 regularization and C=1.0. For the neural method, we use a bi-directional LSTM with three layers of size 900 and a dot attention layer. Loss functions are negative log likelihood, binary cross entropy and mean squared error loss for single-label, multi-label and regression datasets respectively.

Each word in the utterance of interest is represented as a vector with its lexicon scores as values. The dimension of the vector is thus equal to the number of labels in the lexicon. These scores are then averaged over the words to get an average vector for the complete data instance. In some lexica, not all words get scores for every label (see e.g. \textit{Anger}, \textit{Joy} and \textit{Sadness} in \texttt{NRC Affect Intensity} in Table \ref{tab:alien}). In that case, we treat the label as 0.

\subsubsection*{VAE dimensionality}
We determine which dimensionality of the VAE latent space is most appropriate in the emotion detection experiments on the described datasets. Different sizes for the hidden variable are used, motivated by their correspondence to psychological emotion frameworks. We try dimensionalities $N=3, N=6, N=8$ and $N=40$, corresponding to the VAD model, the models of Ekman and Plutchik and the dimension of the naively concatenated lexicon feature vector respectively. We further try dimensions $N=10, N=20$ and $N=30$ to assess whether just adding more features is more predictive than learning a valuable representation of the data. We use a logistic regression classifier for the categorical datasets and linear regression for the continuous datasets.

\subsubsection*{Experiments}
In a first set of experiments, we use the information from each lexicon separately as features in a simple machine learning model to predict labels/scores for each of the thirteen datasets. We use a logistic regression classifier for the categorical datasets and linear regression for the continuous datasets.


Using the same algorithms, we also compute performances for all datasets when the different lexica are combined. We explore three options: using a naive concatenation of all lexica (resulting in a combined feature vector of dimension 40), using the VAE joint lexicon, and using a naive combination where the VAE lexicon is concatenated as an additional lexicon (feature vector with dimension 40 + number of latent dimensions).

We then explore the use of the combined lexica in a neural network approach, using a bi-directional LSTM. Our data is transformed to lexicon vectors (using the naively concatenated representation, the VAE lexicon and the naive concatenation plus VAE lexicon). Each data instance is thus again represented as a feature vector where the words are represented by their lexicon scores. We compare two versions of the network: one where we update the weights (lexicon scores) while training, and one where we keep the weights fixed.

We further test whether lexica can offer complementary gains to neural approaches, which typically rely solely on embeddings. We do this by concatenating the lexicon features in the Bi-LSTM with GloVe word embeddings (200-dimensional embeddings, pre-trained on Twitter data) \citep{pennington2014glove} and the state-of-the-art BERT embeddings \citep{devlin-etal-2019-bert} as input features. Because we are merely interested in comparing different approaches and not in finding the best model per se, we do not perform a large grid search over hyperparameters for our networks. For BERT, we simply use the pretrained BERT model and the PyTorch interface for BERT by Hugging Face \citep{Wolf2019HuggingFacesTS} and use the word vectors of the last layer (768-dimensional embeddings) in the BERT model as input word vectors. We investigate how our Bi-LSTM performs with only word embeddings as features and how the performance alters when lexicon features are added (again with the three scenarios discussed above). We both try fine-tuning the embeddings and keeping the pre-trained embeddings fixed.

We are interested in a) the strengths of individual lexica, for example regarding agreement of framework between lexicon and dataset or the effect of lexicon size and construction method; b) the effect of combining lexica compared to using individual lexica, more specifically when using latent representations from a VAE compared to a naive concatenation of lexica; c) the performance of different machine learning methods (although we limit ourselves to basic approaches and do not heavily tune those) and word representations; d) the performance of using lexica in combination with word embeddings compared to word embeddings or lexica on their own and e) the performance of using fixed lexicon scores compared to trainable inputs.



\subsection{Results}\label{sec:results}
\subsubsection{VAE dimensionality}\label{sec:vae_dim}
First, we determine which dimensionality of the VAE latent space is most appropriate for using it as features to predict emotions in thirteen datasets (logistic regression for classification and linear regression for regression). Different dimensionalities are investigated: $N=\{3,6,8,10,20,30,40\}$. Table \ref{tab:vae} shows the performance of the different VAE versions when used as features in the linear model. Differences in performance seem to be only minor. Indeed, when testing for significance using Welch ANOVA, no significant difference in mean between the performances of different VAE dimensionalities (using performance on separate datasets as data points) is found ($F=0.053, P=0.999$ for single label datasets; $F=0.020, P=1.000$ for multi-label datasets; $F=0.073, P=0.998$ for regression datasets)\footnote{We took the performance on each dataset as data points per VAE dimensionality. For the single-label datasets, we have $8*7$ datapoints (8 datasets and 7 VAE versions); for the multi-label datasets, we take each label as a separate data point, resulting in $19*7$ data points; for regression, we again take each dimension as a separate data point, resulting in $11*7$ data points. We compare means by Welch ANOVA for unequal variances.}. As $N=8$ often leads to the highest results and it was shown we can correlate this dimensionality with Plutchik framework, we choose $N=8$ as final dimensionality.

\begin{table}
\centering
\tiny
\begin{tabular}{l | c | c | c c c c c c c}
\toprule
Dataset	&   Metric  & \#classes &	3-dim	&	6-dim	&	8-dim	&	10-dim & 20-dim & 30-dim & 40-dim	\\
\midrule
\textsc{Blogs}	&	\multirow{8}{*}{Acc.} & 7 & 0.695	&	0.709	&	0.707	&	0.708 & 0.708 & \textbf{0.714} & 0.708	\\
\textsc{Emotion in Text} &	&	13 & 0.271	&	0.280	&	0.286	&	0.290 & \textbf{0.291} & 0.289 & 0.278	\\
\textsc{DailyDialog} &	&	7 & 0.817	&	0.819	&	\textbf{0.820}	&	0.818 & \textbf{0.820} & \textbf{0.820} & 0.819	\\
\textsc{ElectoralTweets} &	&	19 & 0.250	&	0.253	&	\textbf{0.264}	&	0.259 & 0.255 & 0.253 & 0.247	\\
\textsc{Emotion-Stimulus} &	&	7 & 0.536	&	0.646	&	0.644	&	0.636 & 0.604 & \textbf{0.665} & 0.633	\\
\textsc{ISEAR} &	&	7 & 0.271	&	0.378	&	0.374	&	0.372 & 0.388 & \textbf{0.390} & 0.353	\\
\textsc{Tales} &	&	5 & 0.467	&	0.554   &	\textbf{0.574}	&	0.562 & 0.550 & 0.537 & 0.500	\\
\textsc{TEC} &	&	6 & 0.417	&	0.451	&	0.454	&	\textbf{0.466} & 0.458 & 0.450 & 0.437	\\
\hline
\textsc{Affect in Tweets}	&   \multirow{2}{*}{Jacc.}   &	11 & 0.332	&	0.381	&	0.394	&	0.388 & \textbf{0.401} & 0.381 & 0.356	\\
\textsc{SSEC} &	&	8 & 0.425	&	0.434	&	0.432	&	\textbf{0.439} & 0.432 & 0.433 & 0.426	\\
\hline
\textsc{Affective Text}	&   \multirow{3}{*}{$r$}   &	 & 0.308	&	0.336	&	0.323	&	\textbf{0.347} & 0.304 & 0.303 & 0.294	\\
\textsc{EmoBank} &	& &	0.262	&	0.282	&	0.298	&	0.308 & 0.320 & \textbf{0.344} & 0.343	\\
\textsc{Facebook-VA} &	& &	0.382	&	0.384	&	0.385	&	0.386 & \textbf{0.393} & 0.386 & 0.392	\\
\bottomrule
\end{tabular}
\caption{Emotion prediction results per dataset for the VAE lexicon with different dimensions. Accuracy is reported for single-label classification, Jaccard accuracy for multi-label classification and Pearson's $r$ for regression. The best results per dataset are marked in bold.
\newline
See Table \ref{tab:datasets} for an overview of the target labels in each dataset.}
\label{tab:vae}
\end{table}

\subsubsection{Individual lexica}\label{sec:individual}
We train linear and logistic regression classifiers for each dataset separately with lexicon scores as features (see Table \ref{tab:datasets} for an overview of the target labels in each dataset). Table \ref{tab:results_logreg} (upper part) reports the average accuracy, aggregated over the different single-label datasets, average Jaccard accuracy for the multi-label datasets and average Pearson correlation for the regression datasets. The results for individual datasets are shown in Table \ref{tab:results_individual} (Segment 1). First, the individual lexica are used separately, and overall, the \texttt{NRC Hashtag} lexicon is the most predictive one.  More specifically, \texttt{NRC Hashtag} was the best lexicon for nine out of thirteen datasets (\textsc{Emotion in Text}, \textsc{DailyDialog}, \textsc{ElectoralTweets}, \textsc{ISEAR}, \textsc{Tales}, \textsc{Affect In Tweets}, \textsc{SSEC}, \textsc{Affective Text} and \textsc{Facebook-VA}). In three datasets, \texttt{NRC Affect Intensity} is the best lexicon overall. \texttt{Stevenson} gives the best performance on one dataset (\textsc{Tales}).

\texttt{Affective Norms}, \texttt{NRC VAD}, \texttt{ANEW} and \texttt{WordNet Affect} are most often the least predictive lexica. This indicates that lexica with a VAD-framework are less suited for emotion prediction than lexica annotated with (scores for) categories. Moreover, even for the datasets that are annotated with dimensions (\textit{valence} and \textit{arousal} in \textsc{Facebook-VA} and \textit{valence}, \textit{arousal} and \textit{dominance} in \textsc{EmoBank}), \texttt{NRC Hashtag} and \texttt{NRC Affect Intensity} are respectively the best lexica. Although one could suggest that VAD lexica perform better on dimensional datasets than on categorical datasets (with \texttt{Affective Norms} performing second best on \textsc{Facebook-VA} and \texttt{NRC VAD} second best on \textsc{EmoBank}) there is no sign that VAD lexica are more suitable for datasets with dimensional annotations than categorical lexica.

One factor that could influence the performance of the lexicon is the lexicon size. However, we find that this is not at all decisive. The second best performing lexicon is \texttt{NRC Affect Intensity}, but with its 4,192 unique words, this lexicon is rather small. Also \texttt{Stevenson} performs fairly well, although only containing 1,034 words. On the other hand, the largest lexicon is \texttt{NRC VAD}, but this lexicon performs rather badly (probably because it has VAD annotations instead of categorical annotations).

\begin{table}
\centering
\small
	\begin{tabular}{l|c|c|c}
\toprule
 & Single-Label & Multi-Label & Regression \\
 & Accuracy (micro-F1) & Jaccard accuracy & Pearson's $r$ \\
 \midrule
$^\bigtriangleup$ \texttt{NRC Hashtag} &	\textbf{0.468}	&	\textbf{0.361}  &	\textbf{0.268}		\\
$^\bigtriangleup$ \texttt{NRC Affect Intensity} &	0.459	&	0.311   &	0.265		\\
$^\bigtriangleup$ \texttt{WordNet Affect}	&	0.450	&	0.246   &	0.122		\\
$^\bigtriangleup$ \texttt{Stevenson} 	&	0.444	&	0.274   &	0.176		\\
$^\bigtriangleup$ \texttt{NRC Emotion}	&	0.441	&	0.305   &	0.207		\\
$^\bigcirc$ \texttt{Affective Norms} 	&	0.420	&	0.297   &	0.244		\\
$^\bigcirc$ \texttt{NRC VAD} 	&	0.414	&	0.269   &	0.245		\\
$^\bigcirc$ \texttt{ANEW} 	&	0.410	&	0.246   &	0.137		\\
\midrule
combi (-vae)	&	0.539	&	0.415   &	0.321		\\
vae	&	0.515	&	0.413   &	\textbf{0.335}		\\
combi (+vae)	&	\textbf{0.549}	&	\textbf{0.426}  &   0.329		\\
\bottomrule
    \end{tabular}
\caption{Results aggregated over datasets (macro-average) for separate lexica and combinations of lexica with logistic/linear regression.
\newline
$^\bigtriangleup$ categorical lexicon
\newline$
^\bigcirc$ dimensional lexicon
\newline
Combinations of lexica: combi = naive concatenation, vae = combination with VAE lexicon; vae+combi = naive concatenation with VAE lexicon included.
\newline
Single-Label datasets = \textsc{Blogs}, \textsc{Emotion in Text}, \textsc{DailyDialog}, \textsc{ElectoralTweets}, \textsc{Emotion-Stimulus}, \textsc{ISEAR}, \textsc{Tales}, \textsc{TEC}.
\newline
Multi-Label datasets =  \textsc{Affect in Tweets}, \textsc{SSEC}.
\newline
Regression datasets = \textsc{Affective Text}, \textsc{EmoBank}, \textsc{Facebook-VA}.
\newline
See Table \ref{tab:datasets} for an overview of the target labels in each dataset.
}
\label{tab:results_logreg}
\end{table}

\subsubsection{Combining lexica in linear classifiers}\label{sec:combi_linear}
Again using linear and logistic regression classifiers, we test combinations of the different lexica for the emotion analysis tasks. Here we get the chance to evaluate the (8-dimensional) joint VAE lexicon and compare it with a naive concatenation of the original lexica. The results are given in the lower part of Table \ref{tab:results_logreg} (results averaged over datasets) and in Table \ref{tab:results_individual} (results for individual datasets, Segment 2).


Contrary to what we expected, the VAE lexicon performs better for only four datasets compared to the naive concatenation (\textsc{ElectoralTweets}, \textsc{Tales}, \textsc{EmoBank} and \textsc{Facebook-VA}). However, adding the VAE dimensions to the naive concatenation (resulting in a 48-dimensional feature vector), resulted in the best accuracy score for ten out of the thirteen datasets. Table \ref{tab:results_logreg} shows that, on average, this combination approach works best for the single-label and multi-label datasets. This seems to suggest that the VAE lexicon and the original lexica on their own capture complementary information, in the same way that unigram and bigram features can capture different aspects of useful information.

\subsubsection{Combining lexica in a Bi-LSTM}\label{sec:combi_bilstm}
We compare the same three scenarios as in the previous section (naive concatenation, VAE lexicon, and naive concatenation with VAE included), but now we use a neural network with Bi-LSTM layers. Table \ref{tab:results_rnn} (first three rows) shows the results for these experiments, aggregated over the single-label datasets, over the multi-label ones and the regression datasets. Here, only the results where the weights of the sentence vector were updated while training are reported. Results for individual datasets and fixed inputs are reported in Table \ref{tab:results_individual} (Segment 3). Overall, updating the lexicon weights performs better. This might be due to the domain discrepancy between datasets and lexica (even though we combined different lexica). Therefore, we hypothesise that training the VAE jointly with the classification network would perform better. This is something future research will need to confirm.

\begin{table}
\centering
\small
	\begin{tabular}{l|c|c|c}
\toprule
 & Single-Label & Multi-Label & Regression \\
 & Accuracy (micro-F1) & Jaccard accuracy & Pearson's $r$ \\
 \midrule
combi	&	\textbf{0.427}	&	0.166   &	0.107		\\
vae	&	0.405	&	0.181   &	\textbf{0.115}		\\
combi+vae	&	0.421	&	\textbf{0.272}  &	-0.064		\\
 \midrule
GloVe	&	0.580	&	0.432   &	0.259		\\
GloVe+combi	&	\textbf{0.604}	&	\textbf{0.475}  &	0.110		\\
GloVe+vae	&	0.588	&	0.463   &	\textbf{0.274}		\\
GloVe+combi+vae	&	0.595	&	0.442   &	0.232		\\
\midrule
BERT	&	0.644	&	0.512   &	\textbf{0.397}		\\
BERT+combi	&	0.637	&	\textbf{0.538}  &	0.275		\\
BERT+vae	&	\textbf{0.648}	&	0.507   &	0.347		\\
BERT+combi+vae	&	0.643	&	0.499   &	0.370		\\
\bottomrule
    \end{tabular}
\caption{Results aggregated over datasets (macro-average) for combinations of lexica in Bi-LSTM.
\newline
See Table \ref{tab:results_logreg} for datasets and abbreviations.}
\label{tab:results_rnn}
\end{table}

In general, when only lexicon features are used, the linear/logistic regression classifier (see Table \ref{tab:results_logreg}) performs better than the Bi-LSTM, probably because the datasets are rather small and the classifier has (in this setup at least) few features to fit to, which makes it far from optimal for neural network based approaches. We again see that the naive concatenation with the VAE lexicon included works best for the multi-label datasets and the VAE lexicon on its own works best for the regression datasets. However, on average, the naive concatenation works best for the single-label datasets.



\begin{landscape}

\begin{table}
\centering
\tiny
	\begin{tabular}{c|l|cccccccc|cc|ccc}
    \toprule
    	& &	\textsc{Blogs}	&	\textsc{Emotion}	&	\textsc{Daily}	&	\textsc{Elect.}	&	\textsc{Emot.}	&	\textsc{ISEAR}	&	\textsc{Tales}	&	\textsc{TEC}	&	\textsc{AffectIn}	&	\textsc{SSEC}	&	\textsc{Affect.}	&	\textsc{Emo}	&	\textsc{Facebook}	\\
    	&	&	& \textsc{In Text}	&	\textsc{Dialog}	&	\textsc{Tweets}	&	\textsc{Stim.}	&	&	&	&	\textsc{Tweets}	&	&	\textsc{Text}	&   \textsc{Bank}	&	\textsc{VA}	\\
	\midrule
 & & \multicolumn{8}{c}{Accuracy} & \multicolumn{2}{c}{Jaccard accuracy} & \multicolumn{3}{c}{Pearson's $r$}\\
\midrule
\multirow{8}{*}{Segm. 1} & $^\bigtriangleup$ \texttt{NRC Hashtag}	&	0.688	&	\textbf{0.280}	&	\textbf{0.820}	&	\textbf{0.229}	&	0.549	&	\textbf{0.379}	&	0.347	&	\textbf{0.451}	&	\textbf{0.311}	&	\textbf{0.411}	&	\textbf{0.294}	&	0.212	&	\textbf{0.296}	\\
& $^\bigtriangleup$ \texttt{NRC Affect Intensity}	&	\textbf{0.690}	&	0.275	&	0.818	&	0.220	&	\textbf{0.641}	&	0.314	&	0.318	&	0.395	&	0.227	&	0.395	&	0.266	&	\textbf{0.272}	&	0.256	\\
& $^\bigtriangleup$ \texttt{WordNet Affect}	&	0.685	&	0.259	&	0.817	&	0.220	&	0.604	&	0.298	&	0.318	&	0.395	&	0.109	&	0.383	&	0.056	&	0.162	&	0.149	\\
& $^\bigtriangleup$ \texttt{Stevenson}	&	0.688	&	0.252	&	0.818	&	0.225	&	0.414	&	0.329	&	\textbf{0.405}	&	0.424	&	0.184	&	0.364	&	0.160	&	0.138	&	0.231	\\
& $^\bigtriangleup$ \texttt{NRC Emotion}	&	0.689	&	0.272	&	0.817	&	0.220	&	0.520	&	0.290	&	0.335	&	0.387	&	0.235	&	0.374	&	0.175	&	0.218	&	0.226	\\
& $^\bigcirc$ \texttt{Affective Norms}	&	0.685	&	0.263	&	0.816	&	0.217	&	0.398	&	0.208	&	0.376	&	0.398	&	0.205	&	0.390	&	0.248	&	0.202	&	0.280	\\
& $^\bigcirc$ \texttt{NRC VAD}	&	0.684	&	0.252	&	0.816	&	0.220	&	0.401	&	0.196	&	0.335	&	0.406	&	0.164	&	0.374	&	0.279	&	0.224	&	0.230	\\
& $^\bigcirc$ \texttt{ANEW}	&	0.687	&	0.250	&	0.816	&	0.224	&	0.330	&	0.214	&	0.368	&	0.392	&	0.129	&	0.364	&	0.101	&	0.126	&	0.185	\\
\hline																																					
\multirow{3}{*}{Segm. 2.} & combi (-vae)	&	0.718	&	0.308	&	0.821	&	0.255	&	0.678	&	\textbf{0.472}	&	0.562	&	0.501	&	0.394	&	0.436	&	0.296	&	0.328	&	0.339	\\
& vae	&	0.707	&	0.286	&	0.820	&	0.264	&	0.644	&	0.374	&	0.574	&	0.454	&	0.394	&	0.432	&	\textbf{0.323}	&	0.298	&	\textbf{0.385}	\\
& combi (+vae)	&	\textbf{0.726}	&	\textbf{0.315}	&	\textbf{0.822}	&	\textbf{0.278}	&	\textbf{0.683}	&	0.459	&	\textbf{0.603}	&	\textbf{0.506}	&	\textbf{0.412}	&	\textbf{0.440}	&	0.279	&	\textbf{0.347}	&	0.361	\\
\hline																																					
\multirow{6}{*}{Segm. 3} & fxd combi	&	0.677	&	0.249	&	0.816	&	0.221	&	0.267	&	0.224	&	0.285	&	0.404	&	0.018	&	0.282	&	\textbf{0.062}	&	\textbf{0.350}	&	-0.111	\\
& trn combi	&	0.683	&	0.277	&	0.810	&	\textbf{0.270}	&	0.267	&	\textbf{0.341}	&	0.331	&	0.434	&	0.016	&	0.316	&	0.045	&	0.130	&	\textbf{0.145}	\\
& fxd vae	&	0.685	&	0.249	&	0.817	&	0.222	&	0.254	&	0.188	&	0.289	&	0.406	&	0.016	&	0.269	&	0.044	&	0.133	&	-0.010	\\
& trn vae	&	\textbf{0.687}	&	0.272	&	\textbf{0.828}	&	0.198	&	0.289	&	0.206	&	\textbf{0.351}	&	0.409	&	0.019	&	0.342	&	0.035	&	0.196	&	0.113	\\
& fxd combi+vae	&	\textbf{0.687}	&	0.252	&	0.809	&	0.216	&	0.254	&	0.196	&	0.306	&	0.404	&	0.039	&	0.295	&	0.021	&	0.246	&	-0.011	\\
& trn combi+vae	&	0.643	&	\textbf{0.294}	&	0.783	&	0.264	&	\textbf{0.349}	&	0.266	&	0.331	&	\textbf{0.438}	&	\textbf{0.203}	&	\textbf{0.342}	&	0.037	&	-0.307	&	0.079	\\
\hline																																					
\multirow{8}{*}{Segm. 4} & fxd GloVe	&	0.689	&	0.287	&	0.817	&	0.263	&	0.483	&	0.429	&	0.388	&	0.437	&	0.111	&	0.393	&	0.236	&	-0.167	&	0.330	\\
& trn GloVe	&	0.674	&	0.342	&	0.838	&	0.269	&	0.911	&	0.538	&	0.508	&	0.563	&	0.396	&	0.469	&	0.244	&	0.190	&	0.343	\\
& fxd GloVe+combi	&	0.692	&	0.259	&	0.821	&	0.271	&	0.461	&	0.373	&	0.397	&	0.450	&	0.145	&	0.483	&	0.257	&	-0.396	&	0.235	\\
& trn GloVe+combi	&	\textbf{0.748}	&	\textbf{0.360}	&	\textbf{0.850}	&	\textbf{0.295}	&	\textbf{0.948}	&	0.515	&	\textbf{0.541}	&	\textbf{0.577}	&	\textbf{0.463}	&	0.487	&	0.127	&	-0.142	&	0.346	\\
& fxd GloVe+vae	&	0.705	&	0.288	&	0.822	&	0.285	&	0.605	&	0.454	&	0.360	&	0.460	&	0.210	&	0.399	&	0.182	&	-0.178	&	0.040	\\
& trn GloVe+vae	&	0.735	&	0.348	&	0.842	&	0.263	&	0.913	&	\textbf{0.541}	&	0.508	&	0.556	&	0.437	&	\textbf{0.490}	&	0.243	&	\textbf{0.230}	&	0.350	\\
& fxd GloVe+combi+vae	&	0.686	&	0.263	&	0.812	&	0.281	&	0.475	&	0.406	&	0.405	&	0.473	&	0.282	&	0.440	&	0.298	&	0.089	&	0.305	\\
& trn GloVe+combi+vae	&	0.724	&	0.359	&	0.847	&	0.284	&	0.934	&	0.511	&	0.529	&	0.570	&	0.447	&	0.437	&	\textbf{0.316}	&	-0.082	&	\textbf{0.461}	\\
\hline																																					
\multirow{8}{*}{Segm. 5} & fxd BERT	&	0.731	&	0.336	&	0.836	&	0.290	&	0.671	&	0.499	&	0.603	&	0.555	&	0.398	&	0.530	&	\textbf{0.376}	&	-0.200	&	0.611	\\
& trn BERT	&	0.800	&	\textbf{0.389}	&	\textbf{0.854}	&	0.279	&	\textbf{0.857}	&	\textbf{0.634}	&	0.727	&	0.614	&	0.497	&	0.528	&	0.353	&	0.095	&	0.744	\\
& fxd BERT+combi	&	0.737	&	0.350	&	0.835	&	0.322	&	0.651	&	0.505	&	0.707	&	0.583	&	0.394	&	0.525	&	0.324	&	\textbf{0.216}	&	0.646	\\
& trn BERT+combi	&	0.819	&	0.379	&	0.851	&	0.294	&	0.835	&	0.614	&	0.707	&	0.598	&	\textbf{0.530}	&	\textbf{0.546}	&	0.289	&	-0.201	&	0.737	\\
& fxd BERT+vae	&	0.727	&	0.336	&	0.835	&	0.270	&	0.671	&	0.577	&	0.616	&	0.579	&	0.417	&	0.498	&	0.222	&	-0.077	&	0.631	\\
& trn BERT+vae	&	\textbf{0.829}	&	0.381	&	0.850	&	\textbf{0.328}	&	0.851	&	0.629	&	0.698	&	\textbf{0.618}	&	0.499	&	0.514	&	0.236	&	0.053	&	\textbf{0.753}	\\
& fxd BERT+combi+vae	&	0.749	&	0.356	&	0.834	&	0.293	&	0.612	&	0.575	&	0.640	&	0.580	&	0.417	&	0.525	&	0.307	&	-0.112	&	0.629	\\
& trn BERT+combi+vae	&	0.803	&	0.378	&	0.850	&	0.293	&	0.841	&	0.634	&	\textbf{0.736}	&	0.613	&	0.512	&	0.486	&	0.299	&	0.116	&	0.696	\\
    \bottomrule
\end{tabular}
\caption{Results per dataset for separate lexica, combinations of lexica and GloVe/BERT embeddings with logistic/linear regression or Bi-LSTM.
\newline
fxd = fixed embeddings, trn = trainable embeddings. See Table \ref{tab:results_logreg} for other abbreviations.
\newline
\tiny{Most differences in performance are not significant when the datasets are viewed as data points. Significance was tested for difference in performance between the approaches in each segment by taking the performance on each dataset as data points with separate tests for single-label, multi-label and regression datasets, using Kruskal-Wallis H test. Only for the regression datasets in Segment 1, we found a statistically significant difference in performance. See Table \ref{tab:significance} in \ref{appendix:c} for $H-$ and $P$-values and more information about the calculation.}}
\label{tab:results_individual}
\end{table}
\end{landscape}

\subsubsection{GloVe and BERT}\label{sec:glove_bert}
Results aggregated over datasets are shown in Table \ref{tab:results_rnn} (rows 4-11). The results for individual datasets are sown in Table \ref{tab:results_individual} (Segments 4 and 5). We find that, when using GloVe embeddings, adding lexicon information always boosts performance. In most cases (especially for the single-label and regression datasets), adding the naive lexicon concatenation works best, but in some adding the VAE lexicon performs better. Overall, the models with GloVe (strongly) outperform the models with only lexicon features, although models with only GloVe embeddings (without lexicon information) do not perform better than when lexicon information is added.


For BERT, we see a different pattern. Here, adding lexicon information still performs better for the majority of datasets, but not for every single one. In four cases (\textsc{Emotion in Text}, \textsc{DailyDialog}, \textsc{ISEAR}, \textsc{Affective Text}), a model with only BERT embeddings as input performs best. For the other datasets, the best performing combination was often BERT combined with the VAE lexicon. Variants of the BERT model work best for all datasets except for \textsc{Emotion-Stimulus} and \textsc{EmoBank}, where, respectively, trainable GloVe with the naive concatenation and the fixed naive concatenation in a Bi-LSTM work best.
This pattern is in line with findings of related work: state-of-the-art models such as BERT lessen the need for lexicon-based features. However, we show that for the majority of datasets, adding lexicon information still offers additional gains compared to plain embedding models.

\subsubsection{Comparison with benchmark results}\label{sec:sota}

For reference, we report the highest metrics achieved in other studies dealing with the datasets of interest. Table \ref{tab:sota} shows these scores in the metric as reported in the referred study. For two datasets -- \textsc{DailyDialog} and \textsc{Emotion-Stimulus} -- we have not found any benchmark results, as these datasets were originally not developed for the task of emotion detection, but for response retrieval/generation and detection of the causes of emotions respectively. Most of these results are not directly comparable with our results as different train and test splits were used.

\begin{table}
\centering
\scriptsize
	\begin{tabular}{p{2.2cm}l|p{2.3cm}l|p{0.8cm}p{2.5cm}}
\toprule
	&	Metric	&	\multicolumn{2}{c|}{Ours}	&	\multicolumn{2}{c}{SOTA}	\\
	&		&	Model & Score	&	Score	&	Reference	\\
	\midrule
\textsc{Blogs*}	&	Macro-F1	&	BERT+combi+vae$^\triangle$ & 0.616	&	\textbf{0.667}	& \tiny\citet{hosseini2017sentence}		\\
\textsc{Emotion in Text*}	&	Acc.	&	BERT & 0.389	&	\textbf{0.415}**	&	\tiny\citet{li2018generative}	\\
\textsc{DailyDialog}	&	Acc.	&	BERT & 0.854	&	\multicolumn{2}{c}{--}		\\
\textsc{Electoral Tweets*}	&	Acc.	&	BERT+vae & 0.328	&	\textbf{0.568}	&	\tiny\citet{mohammad-2015-electoral}	\\
\textsc{Emotion-Stimulus*}	&	Acc.	&	GloVe+combi & 0.948	&	\multicolumn{2}{c}{--}		\\
\textsc{ISEAR*}	&	Acc.	&	BERT & \textbf{0.634}	&	0.56	&		\tiny\citet{atmajadeep} \\
\textsc{Tales*}	&	Macro-F1	&	BERT+combi+vae$^\triangle$ & \textbf{0.700}	&	0.661	&		\tiny\citet{agrawal-etal-2018-learning} \\
\textsc{TEC*}	&	Macro-F1	&	BERT & \textbf{0.535}	&	0.499	& \tiny\citet{purpura2019supervised}	\\
\hline
\textsc{Affect in Tweets}	&	Jaccard acc.	&   BERT+combi & 0.530	&	\textbf{0.59}	&	\tiny\citet{jabreel2019deep}	\\
\textsc{SSEC}	&	Micro-F1	&	BERT+combi & \textbf{0.691}	&	0.62	&		\tiny\citet{schuff-etal-2017-annotation} \\
\hline
\textsc{Affective Text}	&	Pearson's r	&	BERT$^\triangle$ & 0.376	&	\textbf{0.67}	&	\tiny\citet{buechel2018learning}	\\
\textsc{EmoBank*}	&	Pearson's r	&	combi$^\triangle$ & 0.350	&	\textbf{0.487}***	&	\tiny\citet{wu2019semi}	\\
\textsc{Facebook-VA*}	&	Pearson's r	&	BERT+vae & 0.753	&	\textbf{0.794}***	&	\tiny\citet{wu2019semi}	\\
\bottomrule
\end{tabular}
\caption{Comparison of our best models with state-of-the-art results.
\newline
* No train and test split in original dataset. We used an 80:20 train-test split. 10\% from all data in the training set was set apart for validation. For Blogs, Electoral Tweets and Tales, the studies we compared our results with employed 10-fold cross validation for evaluation, while In TEC, 5-fold cross validation was used. For ISEAR, the reported state-of-the-art metric was obtained in an 80:20 train-test split.
\newline
** Only 50\% of data used, with a 3:1:1 train-dev-test ratio.
\newline
*** Only 40\% of data used for training, 10\% used for testing and 50\% regarded as unlabeled data for semi-supervised learning.
\newline
$^\triangle$Fixed instead of trainable embeddings gave a slightly better performance here.
\newline
-- For \textsc{DailyDialog} and \textsc{Emotion-Stimulus} we have not found any benchmark results, as these datasets were originally not developed for the task of emotion detection, but for response retrieval/generation and detection of emotion \textit{stimuli} respectively.
\newline
See Table \ref{tab:datasets} for an overview of the target labels in each dataset.}
\label{tab:sota}
\end{table}

We find higher results on four datasets, namely \textsc{ISEAR}, \textsc{Tales}, \textsc{TEC} and \textsc{SSEC}, but only \textsc{SSEC} is directly comparable. For all of these four datasets, BERT was the best performing model, although not necessarily with the VAE lexicon. Note that we did not perform a large grid search over hyperparameters for our networks, so it is very likely that our results can be improved further by hyperparameter optimization and fine-tuning BERT representations.

\section{Discussion}\label{sec:discussion}



In this section, we discuss some insights about the use of lexica provided by the emotion detection experiments performed in Section \ref{sec:evaluation}. We zoom in on factors that have a potential impact on the performance of lexica, namely lexicon size, construction method and quality, label set and dimensionality, trainability of the input representation and lexicon combination strategy.

\subsection{Effect of lexicon size}
Different factors play a role in the performance of a lexicon. Vocabulary coverage is a crucial aspect, as lexicon features can only be useful when enough words in the text to be classified have lexicon annotations. Of course, lexicon size and vocabulary coverage are correlated, as the comparison of the individual lexica point out: although the three VAD lexica all perform rather poorly, \texttt{ANEW} is clearly worse than the other two lexica. With only 1,034 words, \texttt{ANEW} has only a limited size and a lot of words in the texts to be classified are not found in the lexicon. On the other hand, the best lexicon, \texttt{NRC Hashtag}, is fairly extended (16,862 words), supporting the hypothesis that large lexica perform better. However, the (regarding label set) rather similar datasets \texttt{Stevenson} and \texttt{NRC Affect Intensity} contain only 1,034 and 4,192 words respectively and also perform reasonably to very well.

The best scores are obtained when all lexica are combined, either as a naively combined lexicon or the joint VAE lexicon. These are by far the largest lexica used in our experiments and indeed, they perform signficantly better than the individual lexica. However, the gain given by combining lexica is much more substantial than the gain of using the approximately 17,000 words in \texttt{NRC Hashtag Emotion Lexicon} compared to the smaller \texttt{Stevenson} or \texttt{NRC Affect Intensity}. This suggests that the benefit of combining lexica not only lies in expanding the vocabulary size, but also in combining the signals coming from various emotion frameworks to build a richer emotion representation for words.

\subsection{Effect of construction method and quality}
It is compelling to link the origin and quality of the lexica to their performance. One may hypothesize that lexica created under lab conditions are of higher quality and thus would be more useful when used as features in machine learning tasks. \texttt{Anew}, \texttt{Stevenson} and \texttt{Affective Norms} were created manually under lab conditions, but show some notable variability among raters: \texttt{Anew} and \texttt{Affective Norms} have ratings from 1 to 7 and have average standard deviations of 1.65, 2.37 and 2.06 for \textit{valence}, \textit{arousal} and \textit{dominance} in \texttt{Anew} and 1.68, 2.30 and 2.16 in \texttt{Affective Norms}; \texttt{Stevenson} has ratings from 1 to 5 for \textit{anger}, \textit{fear}, \textit{sadness}, \textit{joy} and \textit{disgust} and has standard deviations between 0.9 and 1 for all dimensions (standard deviations were calculated per word across the ratings of all raters and then the average was taken per dimension). \texttt{NRC Emotion} and \texttt{NRC Affect Intensity} have been collected through crowdsourcing, which could have an affect on quality as well. \texttt{WordNet Affect} was annotated manually but then automatically expanded with \texttt{WordNet} relations. There is no information about the annotators or quality. Interestingly, the best performing lexicon has been constructed automatically. Lexica created under lab conditions do not necessarily perform well (\texttt{ANEW} performs badly and \texttt{Affective Norms} only average), while crowdsourced lexicon annotations can give fairly good results (as in the case of \texttt{NRC Affect Intensity}).

\subsection{Effect of label set}
We find that lexica with categorical annotations perform better than VAD lexica, on the condition that the categorical lexica have real-valued annotations instead of binary values.

\begin{figure}
    \centering
    \includegraphics[scale=0.6]{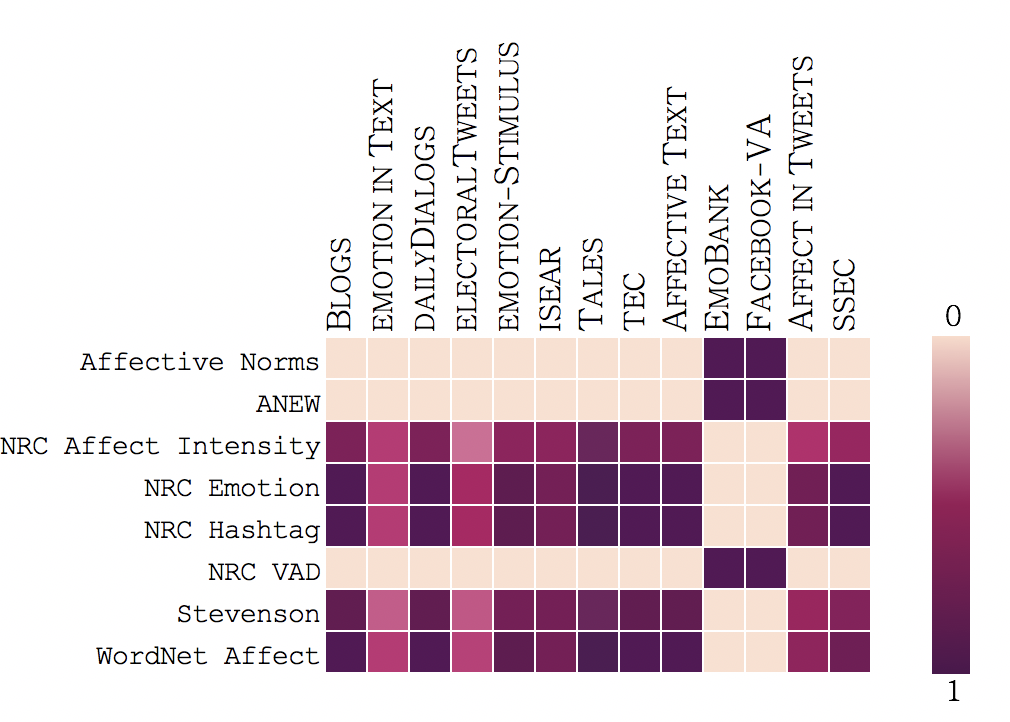}
    \caption{Visualization of overlap between lexica and datasets (number of labels that are shared between the lexicon's and the dataset's label set, normalized by label set size in the dataset). }
    \label{fig:overlap}
\end{figure}

We figured label overlap could play a role in the performance (meaning that the more labels in the lexicon that overlap with the target labels of the dataset, the better the lexicon would perform on that dataset). We define label overlap as the number of labels that are shared between the lexicon's and the dataset's label set, normalized by label set size in the dataset. For example: label overlap between the \texttt{NRC Affect Intensity} lexicon and the \textsc{Electoral Tweets} dataset is 0.21 (\textsc{Electoral Tweets} has 19 labels, of which 4 are shared with the \texttt{NRC Affect Intensity} lexicon). Figure \ref{fig:overlap} visualizes the label overlap between each lexicon and each dataset.

For each dataset, we have eight label overlap scores and eight accuracy scores (one for each lexicon), which allows us to calculate Pearson correlation between the label overlaps and accuracies per dataset (see Table \ref{tab:label_overlap}). For regression datasets, the label overlap hypothesis does not hold: VAD lexica are not better in predicting VAD scores in datasets than categorical lexica are (see Table \ref{tab:results_logreg} in Section \ref{sec:results}). This might be due to the complexity of annotating VAD \citep{Mohammad2018ACL}, resulting in different interpretations of the concepts in the lexica and target datasets and making other -- non-VAD -- lexica more informative. For half of the remaining datasets, there is a high correlation between label overlap and accuracy. This translates into the claim that the more labels are annotated in the lexicon (and thus the higher the chance of a big label overlap), the better the lexicon performs.
 
\begin{table}
\centering
\small
\begin{tabular}{l|c||l|c}
Dataset & $r$ & Dataset & $r$ \\
\toprule
\textsc{Blogs} & 0.243 & \textsc{Affect in Tweets} & 0.642 \\
\textsc{Emotion in Text} & 0.651 & \textsc{SSEC} & 0.326 \\
\textsc{DailyDialog} & 0.724 & \textsc{Affective Text} & -0.179 \\
\textsc{Electoral Tweets} & 0.416 & \textsc{EmoBank} & -0.170 \\
\textsc{Emotion-Stimulus} & 0.736 & \textsc{Facebook-VA} &  0.000 \\
\textsc{ISEAR} & 0.910 \\
\textsc{Tales} & 0.447 \\
\textsc{TEC} & 0.315 \\
\bottomrule
\end{tabular}
\caption{Pearson correlation between overlap and performance per dataset.}
\label{tab:label_overlap}
\end{table}

 

We also had a look at the coefficients in the linear and logistic regression classifiers to get some more intuition into which lexica were most important in the naive concatenation. The coefficients are shown in Table \ref{tab:coefficients} in \ref{appendix:c}. The influence of different lexica is not equal for all datasets. Even per dataset, it does not seem the case that there are particular datasets that have more importance than others, but we rather see particular emotion categories or dimensions that have consistently more weight (\textit{anger} is the most pronounced one). However, we can not draw conclusions for lexica as a whole by studying these coefficients.

\subsection{Effect of training embeddings}
In the neural network approaches, the input representation of our data consists of word embeddings and/or lexicon vectors. The lexicon vectors can be seen as a pre-trained word embedding, which are concatenated or not with the pre-trained GloVe or BERT embeddings. We perform experiments with fixed pre-trained embeddings and investigate updating the learned weights of the words while training the Bi-LSTM model.

When lexicon vectors are used on their own, updating the word vectors increases performance in more than half of the datasets. When combined with GloVe or BERT embeddings, the trainable setting performs better in almost all cases. This means that tailoring the model to a specific dataset is valuable. Moreover, this also suggests that emotions (or the association of words with certain emotions) are not universal, but rather dataset-dependent or domain-specific. Further research where we get more insights on how emotion scores alter across domains is therefore desirable.

\subsection{Effect of lexicon combination strategy}
We consistently test the difference in performance of a naively concatenated lexicon, the joint VAE lexicon, and a naively concatenated lexicon where the VAE lexicon is included.

In the logistic/linear regression approaches, the naive concatenation with the VAE lexicon included performs best on average. This seems to suggest that the VAE space and the original lexica on their own capture complementary information, in the same way that unigram and bigram features can capture different aspects of useful information.

However, in the neural network approaches, the kind of lexicon information that performs best strongly varies over datasets. Combined with GloVe embeddings, it is often the naive concatenation that works best, but combined with BERT, the VAE lexicon results in the best accuracy on several datasets.

While adding lexicon information almost always outperforms the GloVe-embedding-only approach (regardless of the lexicon combination strategy), this observation does not completely hold for BERT. In around half of the cases, the performance of BERT embeddings could not be improved by adding lexicon information, probably because large pre-trained models are already very strong and already encode some kind of emotion information. However, since lexicon information does improve performance in the other cases, we believe employing lexica still has value, especially when there is no access to large pre-trained models like in low-resource languages.

Contrary to what we expected, the VAE lexicon is not unambiguously better than a naive concatenation when used as features in supervised machine learning for emotion detection. We see different explanations for this. Firstly, it could be that the latent emotion representation contains relevant information that is lost during mapping the latent emotion representation from the VAE to the specific emotion framework from the target dataset. Secondly, the conflicting information in the naive concatenation could be less problematic than anticipated. We based this hypothesis on the results of a similar comparison for sentiment lexica, where it was shown that a VAE outperformed a naive concatenation \citep{hoyle-2019-combining}. However, when merging sentiment lexica, only one concept (sentiment polarity) is playing, while in emotion lexica, multiple concepts are quantified (like \textit{valence}, \textit{arousal}, \textit{anger}, \textit{sadness}, etc.). In merging sentiment lexica with a VAE, the concept is thus preserved (it is rather the scales that are unified instead), while in merging emotion lexica with a VAE, some concepts could get underrepresented. This could lead to lower performance than a naive concatenation, where all concepts are preserved (and `conflicting information' could even be useful). However, in other applications of lexica (in psychological experiments or straightforward keyword-based emotion tagging), such conflicting information is unwanted and a unified label set is desired. Therefore, we believe the VAE joint emotion lexicon is still a valuable resource.

\section{Conclusion}\label{sec:conclusion}


This paper addressed the problem of disparate label spaces in emotion lexica and presented an extended, unified lexicon containing 30,273 unique entries. The lexicon was obtained by merging eight existing emotion lexica with a multi-view variational autoencoder. We showed that we can choose the dimension of the VAE latent space so that it is still interpretable, corresponding to emotional dimensions present in the source lexica.

We evaluated the VAE lexicon by using it as features in the downstream task of emotion detection on thirteen datasets and compared it with the use of the  individual source lexica and a naive concatenation thereof, allowing us to explore the impact of different lexicon characteristics like construction approaches, label sets and vocabulary coverage.

We found that lexica with categorical annotations perform better than VAD lexica, on the condition that the categorical lexica have real-valued annotations instead of binary values. Generally, it seems that the more labels are annotated in the lexicon, the better the classification performance on the dataset. In practice, this means that out of the existing emotion lexica, the Plutchik-annotated \texttt{NRC Hashtag Emotion Lexicon} is best. Also in the VAE lexicon, we found that a latent dimension of $N=8$ works best, and linked these dimensions to the Plutchik emotions.

We trained linear and logistic regression classifiers with lexicon scores as features and Bi-LSTMs with GloVe or BERT embeddings and lexicon vectors as input representations. We found that the VAE lexicon outperforms individual lexica, but in contrast to what we expected, the VAE approach is not always convincingly better than the naive concatenation. However, it does contribute to the naive concatenation when added as an extra lexicon, suggesting that it captures complementary information to the individual lexica.

Overall, the BERT model with lexicon features from emotion lexica performs best on average, with the best lexicon combination strategy varying over datasets. This means that emotion lexica can offer complimentary information to even extremely large pre-trained models. Moreover, when large pre-trained models are not available, as in the case of low-resource languages, we believe lexica are especially useful. It would thus be interesting to explore these approaches for low-resource languages in future research.

Although the VAE lexicon performed not unambiguously better as features in a supervised machine learning systems for emotion detection, it does have advantages compared to a naive concatenation of lexica. In contrast to a naive concatenation, the VAE lexicon has a unified label set and can thus be used in other applications of lexica, like in psychological experiments or keyword-based emotion tagging.

\section*{Acknowledgements}
This research was carried out with funding of the Research Foundation - Flanders under a Strategic Basic Research fellowship and supported with a travel grant from Research Foundation - Flanders.

\section*{References}
\bibliography{jelsmfal}

\appendix
\section{}
\label{appendix:a}

\subsection*{Overview of emotion lexica}\vspace{4mm}
\label{sec:lexica}

\subsubsection*{\texttt{ANEW}} The Affective Norms for English Words (\texttt{ANEW}) by \citet{Bradley1999} is the oldest set of normative emotional ratings for English words that is still influential in emotion (analysis) studies. 1,034 words have been rated for \textit{valence}, \textit{arousal} and \textit{dominance} on a 9-point scale. The ratings were obtained under lab conditions and originate from the field of psychology.

\subsubsection*{\texttt{Stevenson}} \citet{Stevenson2007} provide complementary ratings for the words in \texttt{ANEW} on the five discrete emotions \textit{anger}, \textit{fear}, \textit{sadness}, \textit{joy} and \textit{disgust}, on a scale of 1 to 5. Just as \texttt{ANEW}, these ratings were obtained under lab conditions.

\subsubsection*{\texttt{Affective Norms}} With as much as 13,915 lemmas rated for \textit{valence}, \textit{arousal} and \textit{dominance}, the affective norms of \citet{Warriner2013} form a substantial expansion of \texttt{ANEW}. Although originating from the psychology field, ratings were not obtained in the lab, but through crowdsourcing with Amazon Mechanical Turk.

\subsubsection*{\texttt{WordNet Affect}} This resource was created by NLP researchers. \citet{strapparava2004} developed an extension of WordNet \citep{miller1995wordnet} by manually assigning affective labels to a subset of WordNet synsets, containing information about emotions, moods, attitudes, etc.  This set was then expanded by marking WordNet synonyms as having the same emotion. In the SemEval-2007 Affective Text task, \citet{strapparava-mihalcea-2007-semeval} extracted a list of words relevant to the Ekman emotions from \texttt{WordNet Affect}. This list contains 1,116 words with binary association scores (0 or 1) for the Ekman emotions.

\subsubsection*{\texttt{NRC Emotion}} These ratings were created by \citet{Mohammad2013}, specifically with the aim of using them in an NLP context. The ratings were obtained by calling in the crowd, which resulted in 14,182 words annotated with one or multiple of the Plutchik emotions (binary assocation scores).

\subsubsection*{\texttt{NRC VAD}} This lexicon is another crowdsourced resource for emotion analysis in NLP. \citet{Mohammad2018ACL} obtained ratings for \textit{valence}, \textit{arousal} and \textit{dominance} for 20,007 words, resulting in the largest emotion lexicon that is openly available.

\subsubsection*{\texttt{NRC Affect Intensity}} \citet{Mohammad2018LREC} also provide a lexicon with ratings for (the intensity of) the emotions \textit{anger}, \textit{fear}, \textit{sadness} and \textit{joy}. The lexicon contains 4,192 unique words, where each word gets a rating between 0 and 1 for one or more of the emotion categories. This was again a result of a crowdsourcing effort.

\subsubsection*{\texttt{NRC Hashtag Emotion}} Unlike the previously discussed lexica, which were manually created, this lexicon was constructed automatically by computing the strength of association between a word and an emotion (based on the \textsc{Hashtag Emotion Corpus}) \citep{Mohammad2015hashtags}. The lexicon contains real-valued scores for the eight Plutchik emotions for 16,862 words.

\section{}
\label{appendix:b}

\subsection*{Overview of emotion datasets}\vspace{4mm}
\label{sec:datasets}

\subsubsection*{\textsc{Blogs}} This is one of the oldest emotion datasets used in NLP. It was created by \citet{aman-2007-blogs}. They labeled 5,025 sentences (single-label) from blogs with the Ekman emotions, with additional labels \textit{mixed emotion} and \textit{no emotion}. Supplementary information like emotion intensity (low, medium, high) and emotion markers was given as well, but we will only use the single-label emotion information. Moreover, following other studies, we will only use the sentences with high agreement, resulting in a dataset with 4,090 sentences where the \textit{mixed emotion} category is discarded.

\subsubsection*{\textsc{Emotion in Text}} This dataset was published by CrowdFlower (currently known as Figure Eight), an online data annotation platform. Crowdsourced annotations were collected for 40,000 tweets on the emotion categories \textit{anger}, \textit{boredom}, \textit{empty}, \textit{enthusiasm}, \textit{fun}, \textit{happiness}, \textit{hate}, \textit{love}, \textit{relief}, \textit{sadness}, \textit{surprise}, \textit{worry} and a \textit{neutral} category (single-label).

\subsubsection*{\textsc{DailyDialog}} This fairly recent dataset was published by \citet{Li2017} and consists of 13,118 sentences from dialogs. The dataset is developed for the task of response retrieval and generation, but additionally, emotion information was annotated. The sentences are labeled following the Ekman emotions (with an additional \textit{no emotion} label) in a single-label manner.

\subsubsection*{\textsc{ElectoralTweets}} \citet{mohammad-2015-electoral} collected 4,058 tweets in the political domain, more specifically with the aim to analyse how public sentiment is shaped when it comes to elections. 4,058 tweets were annotated via crowdsourcing for the categories \textit{acceptance}, \textit{admiration}, \textit{amazement}, \textit{anger} (including \textit{annoyance}, \textit{hostility} and \textit{fury}), \textit{anticipation} (including \textit{expectancy} and \textit{interest}), \textit{calmness} (or \textit{serenity}), \textit{disappointment}, \textit{disgust}, \textit{dislike}, \textit{fear} (including \textit{apprehension}, \textit{panic} and \textit{terror}), \textit{hate}, \textit{indifference}, \textit{joy} (including \textit{happiness} and \textit{elation}), \textit{like}, \textit{sadness} (including \textit{gloominess}, \textit{grief} and \textit{sorrow}), \textit{surprise}, \textit{trust}, \textit{uncertainty} (or \textit{indecision}, \textit{confusion}) and \textit{vigilance}. The annotations are single-label.

\subsubsection*{\textsc{Emotion-Stimulus}} Originally, the purpose of this dataset was to identify emotion causes in texts. However, these data can also be used as an emotion detection dataset, as it contains emotion labels for 2,414 sentences. The annotations are done in a single-label manner with the Ekman categories and the additional category \textit{shame} as labels.

\subsubsection*{\textsc{ISEAR}} In the International Survey on Emotion Antecedents and Reactions, \citet{scherer1994} asked people to report on emotional events for the seven emotions \textit{anger}, \textit{disgust}, \textit{fear}, \textit{guilt}, \textit{joy}, \textit{sadness} and \textit{shame}. The sentences from these reports were extracted and linked to the emotion of interest, resulting in a dataset of 7,665 sentences with one out of seven labels.

\subsubsection*{\textsc{Tales}} Although being the oldest emotion dataset in the NLP field, this dataset from \citet{alm-etal-2005-emotions} is still a popular resource. The full dataset consists of 15,302 sentences from 185 fairy tales, annotated with the Ekman emotions, where the \textit{surprise} category is broken up into \textit{positive surprise} and \textit{negative surprise} and a \textit{neutral} label is added as well. The annotation happened in a single-label way. However, the `high-agreement' version of this dataset, where \textit{anger} and \textit{disgust} are merged, no distinction is made between the kinds of \textit{surprise} and \textit{neutral} sentences are ignored, is used more frequently. We will therefore also rely on this reduced dataset, which comprises 1,207 sentences.

\subsubsection*{\textsc{TEC}} The Twitter Emotion Corpus or \textsc{TEC} was automatically created by \citet{mohammad-2012-emotional} via distant supervision. Emotion word hashtags were used to collect tweets, and the hashtags were used for self-labeling. This resulted in a set of 21,051 tweets with (single-label) Ekman tags.

\subsubsection*{\textsc{Affect in Tweets}} In contrast to the previous datasets, the instances in the \textsc{Affect in Tweets} dataset can have multiple labels. The annotations were obtained via crowdsourcing for the Plutchik emotions and three additional labels \textit{love}, \textit{optimism} and \textit{pessimism}. The dataset was used for one of the subtasks in SemEval-2018: Affect in Tweets \citep{mohammad-etal-2018-semeval}.

\subsubsection*{\textsc{SSEC}} The Stance Sentiment Emotion Corpus is another multi-label dataset, published by \citet{schuff-etal-2017-annotation}. It is an extension of the stance and sentiment dataset from SemEval-2016 \citep{mohammad-2016-semeval} and has annotations for the Plutchik emotions for 4,868 tweets.

\subsubsection*{\textsc{Affective Text}} While the aforementioned datasets all contain discrete labels and are intended for emotion classification, the \textsc{Affective Text} dataset from SemEval-2007 by \citet{strapparava-mihalcea-2007-semeval} can be used in regression tasks. 1,250 news headlines were scored for Ekman emotions on a 0 to 100 scale.

\subsubsection*{\textsc{EmoBank}} This dataset by \citet{Buechel2017} is also intended for emotion regression tasks. 10,548 sentences were annotated for the dimensions \textit{valence}, \textit{arousal} and \textit{dominance} on a scale from 1 to 5. The sentences originate from various genres and domains, including the sentences from \textsc{Affective Text} and subsets (blogs, essays, fiction, travel guides, ...) of the Manually Annotated Sub-Corpus of the American National Corpus \cite{ide-2010-manually}.

\subsubsection*{\textsc{Facebook-VA}} \citet{Preoctiuc2016} published a dataset consisting of 2,895 Facebook posts. The posts are annotated for the dimensions \textit{valence} and \textit{arousal} on a 9-point scale and thus are intended for regression tasks.

\section{}
\label{appendix:c}

\subsection*{Supplementary tables}

Table \ref{tab:significance} shows $H-$ and $P$-values of Kruskal-Wallis H tests for testing significance for difference in performance between different emotion detection approaches. The performances on each dataset were taken as data points for each approach. For the multi-label and regression datasets, the performance for each dimension is taken as a separate data point.

Table \ref{tab:coefficients} shows the coefficients of the linear and logistic regression classifiers, this in order to get more intuition into which lexica were most important in the naive concatenation.

\begin{table}[H]
\centering
\tiny
\begin{tabular}{ll|ccc||ll|ccc}
\toprule
\multicolumn{2}{c|}{Segment} & $H$ & $df$ & $P$ & \multicolumn{2}{c|}{Segment} & $H$ & $df$ & $P$ \\
\midrule
\multirow{3}{*}{Segm. 1} & SL & 1.791 & 7 & 0.970 & \multirow{3}{*}{Segm. 4} & SL & 0.315 & 3 & 0.957 \\
 & ML & 2.71 & 7 & 0.910 & & ML & 0.054 & 3 & 0.997 \\
 & Reg & 17.044 & 7 & \textbf{0.017} & & Reg & 3.305 & 3 & 0.347\\
 \midrule
\multirow{3}{*}{Segm. 2} & SL & 0.32 & 2 & 0.852 & \multirow{3}{*}{Segm. 5} & SL & 0.185 & 3 & 0.980 \\
 & ML & 0.09 & 2 & 0.956 & & ML & 0.512 & 3 & 0.916 \\
 & Reg & 0.041 & 2 & 0.980 & & Reg & 0.598 & 3 & 0.897 \\
 \midrule
\multirow{3}{*}{Segm. 3} & SL & 0.099 & 2 & 0.952 \\
 & ML & 0.274 & 2 & 0.872 & \\
 & Reg & 0.683 & 2 & 0.711 & \\
\midrule
\bottomrule
\end{tabular}
\caption{$H$-values, degrees of freedom and $P$-values of Kruskal-Wallis H tests for testing significance for difference in performance between different approaches within segments (see Table \ref{tab:results_individual}).}
\label{tab:significance}
\end{table}

\begin{landscape}

\begin{table}
\centering
\tiny
	\begin{tabular}{l|l|c|c|c|c|c|c|c|c|c|c|c|c|c}
    \toprule
    	Lexicon & Emotional & \textsc{Blogs}	&	\textsc{Emotion}	&	\textsc{Daily}	&	\textsc{Elect.}	&	\textsc{Emot.}	&	\textsc{ISEAR}	&	\textsc{Tales}	&	\textsc{TEC}	&	\textsc{AffectIn}	&	\textsc{SSEC}	&	\textsc{Affect.}	&	\textsc{Emo}	&	\textsc{Facebook}	\\
    	&	cat./dim. &	\textsc{In Text}	&	\textsc{Dialog}	&	\textsc{Tweets}	&	\textsc{Stim.}	&	&	&	&	\textsc{Tweets}	&	&	\textsc{Text}	&   \textsc{Bank}	&	\textsc{VA}	\\
	\midrule
\texttt{nrchashtag}	&	anger	&	4.676	&	1.189	&	2.492	&	-0.581	&	5.071	&	-0.238	&	1.937	&	11.308	&	-0.714	&	0.822	&	-0.159	&	-0.121	&	-3.204	\\
\texttt{nrchashtag}	&	anticipation	&	-0.946	&	-0.911	&	2.745	&	-0.502	&	0.355	&	-0.203	&	0.057	&	-1.428	&	0.143	&	-0.752	&	-7.347	&	0.040	&	-0.481	\\
\texttt{nrchashtag}	&	disgust	&	0.856	&	0.231	&	1.916	&	-0.832	&	-0.416	&	-0.259	&	0.062	&	-0.730	&	-0.289	&	1.152	&	21.539	&	-0.166	&	-2.417	\\
\texttt{nrchashtag}	&	fear	&	-0.805	&	-0.289	&	4.309	&	-0.193	&	-0.590	&	-0.203	&	-1.309	&	-4.102	&	-0.175	&	0.074	&	24.605	&	-0.180	&	-0.168	\\
\texttt{nrchashtag}	&	joy	&	-1.095	&	-0.966	&	-5.290	&	-0.326	&	-3.031	&	-0.163	&	-2.199	&	-6.015	&	-0.379	&	-0.702	&	26.894	&	0.118	&	1.428	\\
\texttt{nrchashtag}	&	sadness	&	-0.695	&	-1.150	&	-0.546	&	-0.098	&	-2.705	&	-0.160	&	-1.717	&	-1.099	&	-0.410	&	0.142	&	8.219	&	-0.340	&	-3.552	\\
\texttt{nrchashtag}	&	surprise	&	-0.818	&	0.054	&	-2.768	&	-0.495	&	-0.142	&	-0.024	&	-0.715	&	-2.740	&	0.121	&	-0.499	&	7.010	&	-0.131	&	-2.190	\\
\texttt{nrchashtag}	&	trust	&	-0.294	&	-0.286	&	-1.987	&	-0.081	&	0.338	&	-0.132	&	0.549	&	1.740	&	0.545	&	0.201	&	-6.679	&	0.148	&	-1.751	\\
\midrule
\texttt{nrcaffect}	&	anger	&	1.141	&	-0.144	&	-3.700	&	-0.055	&	3.890	&	-0.041	&	1.095	&	3.111	&	-0.477	&	0.294	&	60.006	&	0.643	&	6.871	\\
\texttt{nrcaffect}	&	fear	&	-0.678	&	-0.088	&	-0.648	&	-0.238	&	-1.061	&	-0.054	&	-0.108	&	0.087	&	0.162	&	0.306	&	48.413	&	0.070	&	-7.338	\\
\texttt{nrcaffect}	&	joy	&	-0.884	&	-0.235	&	-2.355	&	0.184	&	-2.073	&	-0.071	&	-1.036	&	-2.083	&	-0.204	&	-0.177	&	9.285	&	0.908	&	4.694	\\
\texttt{nrcaffect}	&	sadness	&	-0.270	&	-0.234	&	-2.295	&	-0.083	&	-2.485	&	-0.079	&	-0.392	&	-0.627	&	-0.286	&	0.226	&	48.177	&	-0.726	&	-0.192	\\
\midrule
\texttt{wordnetaffect}	&	anger	&	1.625	&	-0.143	&	-2.010	&	0.064	&	6.940	&	-0.009	&	1.158	&	2.453	&	-0.192	&	0.073	&	0.000	&	-0.096	&	-3.532	\\
\texttt{wordnetaffect}	&	disgust	&	-0.193	&	0.212	&	-0.919	&	-0.045	&	-0.361	&	-0.007	&	0.105	&	0.031	&	0.094	&	-0.069	&	64.406	&	-0.094	&	-1.044	\\
\texttt{wordnetaffect}	&	fear	&	-0.911	&	-0.108	&	-3.958	&	0.014	&	-0.264	&	-0.007	&	-0.897	&	-1.328	&	0.137	&	0.109	&	10.673	&	-0.385	&	-2.137	\\
\texttt{wordnetaffect}	&	joy	&	-0.352	&	-0.076	&	-2.265	&	0.036	&	-1.645	&	-0.030	&	-0.843	&	-0.357	&	-0.562	&	0.079	&	-7.936	&	0.490	&	0.318	\\
\texttt{wordnetaffect}	&	sadness	&	0.029	&	-0.146	&	-2.870	&	0.014	&	-2.931	&	-0.024	&	-0.940	&	1.222	&	-0.085	&	0.056	&	2.903	&	-0.435	&	-0.581	\\
\texttt{wordnetaffect}	&	surprise	&	0.368	&	0.174	&	-2.493	&	-0.033	&	-0.949	&	-0.008	&	0.378	&	0.147	&	0.155	&	-0.127	&	337.291	&	0.625	&	2.142	\\
\midrule
\texttt{stevenson}	&	happiness	&	-1.092	&	-0.695	&	-1.727	&	-0.064	&	-0.691	&	-0.445	&	-1.557	&	-0.517	&	-0.219	&	0.085	&	-15.913	&	-0.020	&	-1.081	\\
\texttt{stevenson}	&	anger	&	1.307	&	0.167	&	0.940	&	-0.508	&	1.844	&	-0.295	&	1.788	&	2.459	&	-0.148	&	0.060	&	-2.709	&	-0.203	&	-0.992	\\
\texttt{stevenson}	&	sadness	&	-0.854	&	-0.092	&	0.912	&	-0.531	&	-1.512	&	-0.311	&	-0.589	&	-0.957	&	-0.201	&	-0.009	&	-14.954	&	0.119	&	0.066	\\
\texttt{stevenson}	&	fear	&	-0.822	&	0.478	&	-0.950	&	-0.390	&	-0.899	&	-0.310	&	-0.406	&	-1.039	&	0.035	&	0.221	&	11.274	&	-0.221	&	-0.576	\\
\texttt{stevenson}	&	disgust	&	-0.077	&	0.233	&	-0.424	&	-0.386	&	-0.204	&	-0.248	&	1.198	&	-2.192	&	-0.071	&	-0.122	&	17.204	&	0.139	&	2.084	\\
\midrule
\texttt{nrcemotion}	&	anger	&	1.151	&	-0.063	&	0.553	&	0.103	&	4.661	&	-0.082	&	1.231	&	1.366	&	-0.644	&	0.426	&	-40.169	&	-0.372	&	-2.967	\\
\texttt{nrcemotion}	&	anticipation	&	-0.688	&	-0.278	&	-0.842	&	-0.170	&	-0.493	&	-0.102	&	-0.587	&	-1.429	&	-0.093	&	-0.038	&	3.405	&	0.004	&	-0.188	\\
\texttt{nrcemotion}	&	disgust	&	0.437	&	-0.013	&	-0.690	&	-0.435	&	0.913	&	-0.069	&	1.089	&	-1.198	&	-0.111	&	-0.055	&	0.266	&	0.224	&	-2.080	\\
\texttt{nrcemotion}	&	fear	&	-0.876	&	-0.205	&	-0.219	&	-0.261	&	-1.276	&	-0.096	&	0.092	&	0.162	&	-0.053	&	0.320	&	-12.310	&	-0.076	&	2.435	\\
\texttt{nrcemotion}	&	joy	&	-0.917	&	-0.397	&	-3.104	&	0.112	&	-1.762	&	-0.077	&	-0.893	&	-0.635	&	-0.509	&	-0.106	&	0.292	&	0.265	&	-1.384	\\
\texttt{nrcemotion}	&	sadness	&	-0.144	&	-0.208	&	-0.192	&	-0.086	&	-1.236	&	-0.113	&	-0.126	&	0.334	&	-0.414	&	0.438	&	-20.880	&	-0.042	&	-2.079	\\
\texttt{nrcemotion}	&	surprise	&	-0.560	&	-0.195	&	0.927	&	0.110	&	-1.840	&	-0.046	&	-0.922	&	-0.184	&	-0.156	&	-0.004	&	-3.125	&	-0.022	&	4.592	\\
\texttt{nrcemotion}	&	trust	&	-0.478	&	0.070	&	2.923	&	0.026	&	0.489	&	-0.135	&	-0.269	&	0.859	&	0.062	&	0.053	&	1.593	&	-0.174	&	0.933	\\
\midrule
\texttt{affectivenorms}	&	valence	&	-0.908	&	-0.824	&	-1.353	&	0.374	&	-1.194	&	-0.116	&	-1.391	&	-0.830	&	0.410	&	-0.463	&	-3.247	&	0.075	&	0.819	\\
\texttt{affectivenorms}	&	arousal	&	1.717	&	0.737	&	0.445	&	-1.220	&	-0.184	&	-1.087	&	0.542	&	0.030	&	-0.210	&	0.411	&	2.239	&	-0.021	&	-0.080	\\
\texttt{affectivenorms}	&	dominance	&	-0.294	&	-0.198	&	1.236	&	0.395	&	1.149	&	0.271	&	1.073	&	0.806	&	-0.325	&	0.229	&	1.418	&	-0.068	&	-0.847	\\
\midrule
\texttt{nrcvad}	&	valence	&	-1.330	&	-0.045	&	-1.729	&	0.373	&	-1.006	&	-0.250	&	-1.080	&	-2.817	&	0.287	&	-0.414	&	9.758	&	0.156	&	-0.944	\\
\texttt{nrcvad}	&	arousal	&	0.575	&	0.697	&	-2.563	&	-0.065	&	0.866	&	-0.418	&	0.396	&	0.988	&	-0.123	&	0.273	&	10.473	&	-0.137	&	4.428	\\
\texttt{nrcvad}	&	dominance	&	-0.267	&	0.550	&	3.861	&	0.591	&	0.837	&	-0.352	&	0.029	&	2.216	&	0.321	&	0.070	&	-35.464	&	0.119	&	-3.249	\\
\midrule
\texttt{anew}	&	valence	&	-0.046	&	0.075	&	1.488	&	0.334	&	-1.107	&	-0.896	&	-1.343	&	-0.232	&	0.310	&	0.145	&	16.234	&	-0.036	&	0.018	\\
\texttt{anew}	&	arousal	&	-0.441	&	-0.105	&	-1.402	&	-0.439	&	-0.417	&	-0.823	&	-0.197	&	0.710	&	-0.091	&	0.075	&	-6.186	&	0.128	&	0.376	\\
\texttt{anew}	&	dominance	&	1.311	&	-0.002	&	0.459	&	0.249	&	1.963	&	-0.826	&	1.103	&	0.329	&	-0.179	&	-0.246	&	-7.020	&	-0.024	&	0.030	\\
\bottomrule
\end{tabular}
\caption{Coefficients of the logistic and linear classifiers per dataset (naive concatenation as input features).}
\label{tab:coefficients}
\end{table}
\end{landscape}

\end{document}